\documentclass[letterpaper, 10 pt, conference]{ieeeconf}
\IEEEoverridecommandlockouts
\usepackage{graphicx}
\usepackage{cite}
\usepackage{picinpar}
\usepackage{amsmath}
\usepackage{url}
\usepackage{flushend}
\usepackage[utf8]{inputenc}
\usepackage{soul}
\usepackage{multirow}
\usepackage{pifont}
\usepackage{color}
\usepackage{alltt}
\usepackage[hidelinks]{hyperref}
\usepackage{enumerate}
\usepackage{siunitx}
\usepackage{epstopdf}
\usepackage{pbox}

\usepackage{epsfig}
\usepackage{mathptmx}
\usepackage{times}
\usepackage{amsmath}
\usepackage{amssymb}

\usepackage{booktabs}
\usepackage{caption}
\usepackage{stfloats}
\usepackage{cuted}
\usepackage{placeins}
\usepackage{cite}  

\usepackage[skins,breakable]{tcolorbox}
\usepackage[percent]{overpic}
\usepackage[caption=false,font=normalsize,labelfont=sf,textfont=sf]{subfig}

\begin{document}
\title{Human-Human \& Human-Robot Interaction Transformer (H$^2$INT) for Robot Navigation in Dense and Uncertain Crowds}

\author{Ao Shen, Kaixi Chen, Shiwei Liu, Fang Deng,~\IEEEmembership{Fellow,~IEEE}, and Chen Chen,~\IEEEmembership{Member,~IEEE}
\thanks{This work was supported in part by the National Science and Technology Major Project under Grant 2022ZD0119703, the National Natural Science Foundation of China under Grant 62273044, and BIT Kunpeng\&Ascend Center of Cultivation. (Corresponding author: Chen Chen.)}
\thanks{Ao Shen, Kaixi Chen, Shiwei Liu, Fang Deng, and Chen Chen are with the School of Automation, Beijing Institute of Technology, Beijing 100081, China, and also with the State Key Laboratory of Autonomous Intelligent Unmanned Systems, Beijing 100081, China (e-mail: 3120235667@bit.edu.cn, 3120250914@bit.edu.cn, 1120222212@bit.edu.cn, dengfang@bit.edu.cn, xiaofan@bit.edu.cn).}}

\maketitle
	
\begin{abstract}
Safe robot navigation in dense crowds requires reasoning about pedestrian motion and how it may change in response to a robot.
However, many learning-based approaches generate pedestrian motion independently of the robot or assume uniform reciprocity, omitting an important source of interaction uncertainty.
This paper presents a Human-Human \& Human-Robot Interaction Transformer (H$^2$INT), a reinforcement learning framework that retains robot-conditioned changes in pedestrian motion during policy learning while allowing responsiveness to vary across pedestrians.
Responsiveness affects the crowd dynamics when the robot is visible but is not supplied as a policy input; the policy must instead infer its consequences from robot-centered relative positions.
A two-stage gated Transformer progressively encodes human-human and human-robot relations, while a recurrent policy captures their temporal evolution.
A curriculum gradually reduces pedestrian responsiveness to increase interaction difficulty.
Simulation experiments demonstrate improved navigation safety and robustness over representative baselines across response conditions and crowd densities, and show transfer without retraining to structurally distinct crowd-flow layouts.
Ablations support the hierarchical relational encoding and gated updates.
Real-robot deployment further verifies that the learned policy can operate with sparse observations in a physical environment.
\end{abstract}


\definecolor{limegreen}{rgb}{0.2, 0.8, 0.2}
\definecolor{forestgreen}{rgb}{0.13, 0.55, 0.13}
\definecolor{greenhtml}{rgb}{0.0, 0.5, 0.0}

\section{Introduction}

\PARstart{S}{ervice} robots are increasingly deployed in restaurants, hospitals, shopping malls, and other human-populated environments~\cite{pieska2013social, silva2023online, gonzalez2021social, shiomi2009field}. In these settings, safe and efficient navigation depends on more than predicting where pedestrians will move. A robot entering a shared path may cause a pedestrian to slow down, yield, or deviate, while another pedestrian may continue without reacting. The robot must therefore act under a feedback loop in which its motion can change pedestrian behavior and the resulting pedestrian motion, in turn, changes the robot's next decision. This behavioral feedback is difficult to exploit because pedestrian motion is uncertain, locally interactive, and only partially observable~\cite{kruse2013human,trautman2015robot, mavrogiannis2023core}.

Existing social-navigation research has made substantial progress in trajectory prediction, graph-based interaction reasoning, and human-aware planning. Some studies explicitly use gaze or body orientation to estimate whether a pedestrian perceives the robot~\cite{ratsamee2013social,ferrer2013robot}. However, many learning-based navigation environments still treat the crowd as an external dynamic obstacle field, assume a fixed reciprocal response, or generate pedestrian trajectories independently of the robot~\cite{ngo2022socially, truong2017socially, 10561574}. These assumptions remove an important source of interaction variability before the policy is trained. A stronger predictor or relational encoder cannot recover feedback that is absent from the environment dynamics; consequently, a policy may perform well under its training assumption yet fail when pedestrians respond differently to the robot.

\begin{figure}[!t]
\centering
\includegraphics[width=0.48\textwidth]{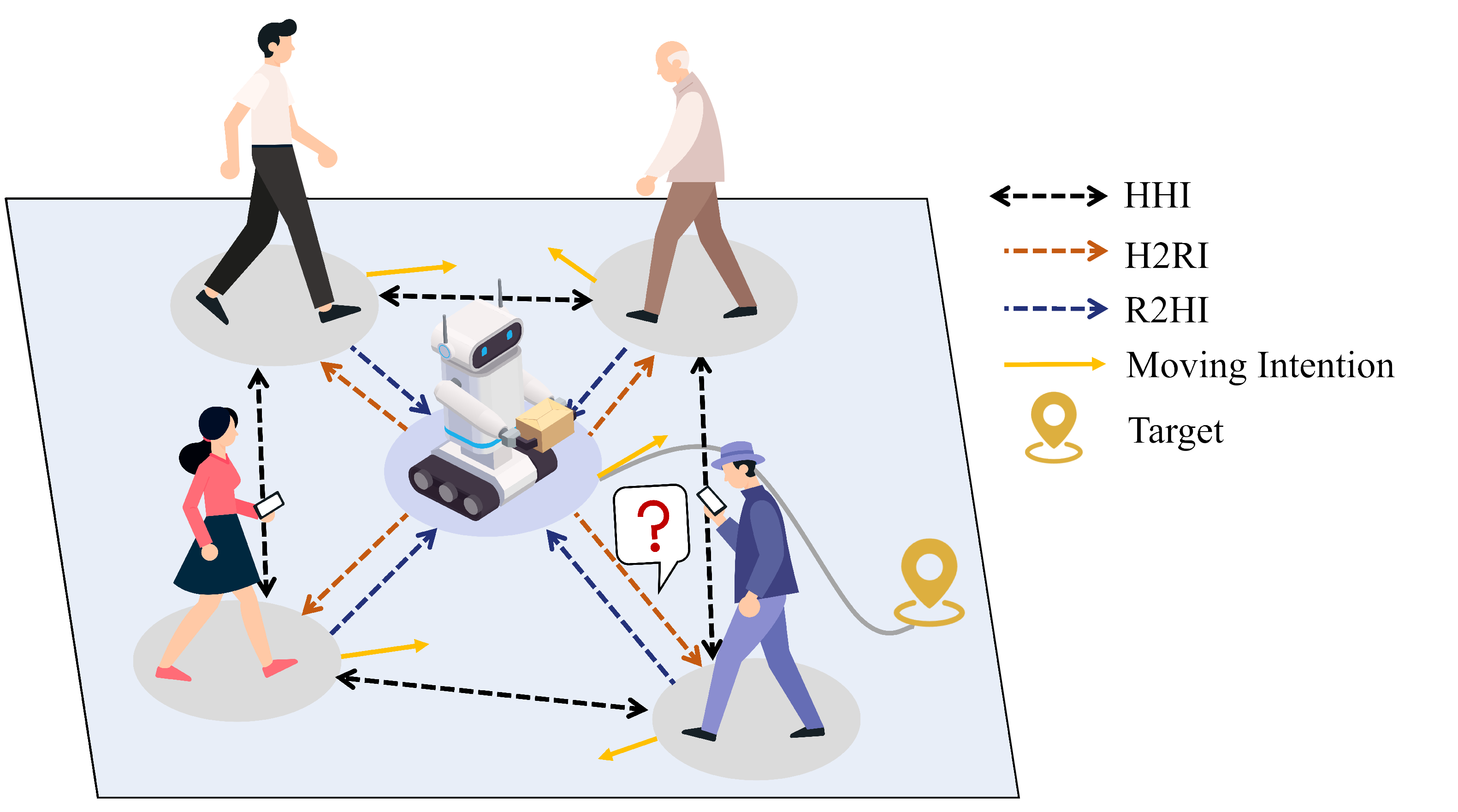}
\caption{An illustration of interaction relationships in crowd navigation.}
\label{fig:wide_example}
\end{figure}

This work formulates the problem at the behavioral rather than cognitive level. Instead of estimating an explicit human attention label, the crowd dynamics allow a geometrically visible pedestrian either to account for the robot during motion generation or to continue independently. The response tendency varies across pedestrians and its binary realization is temporally persistent, avoiding the assumption that every pedestrian reacts identically at every instant. This state remains latent: the robot observes only relative positions over time and must infer the resulting interaction pattern. The simulator thereby exposes the policy to a closed motion-feedback loop without treating cognitive attention as a policy input or requiring noisy pedestrian velocity estimates.

Learning from these sparse observations creates a second challenge: the policy must recover temporal behavior while resolving simultaneous human-human and human-robot dependencies in a variable-size crowd. Since Transformer encoders are already established in social navigation, self-attention alone is not claimed as the contribution. H$^2$INT instead uses two gated Transformer stages to progressively refine the robot-centered interaction representation. GRU-style residual gates regulate information updates within each stage, while a recurrent policy core integrates the refined representation over time.

Building upon these considerations, a Human-Human \& Human-Robot Interaction Transformer (H$^2$INT) is presented for robot navigation in dense and uncertain crowds. The main contributions are summarized as follows:
\begin{itemize}
    \item A behavior-level crowd formulation closes the interaction loop between robot motion and pedestrian response. It accommodates both robot-conditioned and robot-independent pedestrian motion with heterogeneous, temporally persistent response tendencies, while keeping the underlying response state latent to the policy.

    \item A two-stage gated Transformer policy encoder progressively refines human-human and human-robot interaction features from relative positions alone. Gated residual updates regulate information flow through the attention and feed-forward sublayers, and recurrent context supports implicit motion inference under partial observability.

    \item A curriculum progressively reduces pedestrian cooperation during training. Evaluation extends the standard CrowdNav Circle Crossing setting with three structurally distinct test-only flow modes and uses a LiDAR-based real robot to examine geometric transfer and deployment feasibility.

\end{itemize}

\section{Related Work}

Classical reaction-based navigation methods, such as the Social Force Model~\cite{helbing1995social}, Reciprocal Velocity Obstacles~\cite{van2008reciprocal}, ORCA~\cite{van2011reciprocal}, and DWA~\cite{fox2002dynamic}, compute robot actions from current observations using geometric or rule-based constraints. These methods are efficient and interpretable, but they usually rely on symmetric interaction or complete observability assumptions. As a result, they can be overly conservative in dense crowds and may suffer from freezing behavior when many pedestrians interact simultaneously~\cite{trautman2010unfreezing}.

Prediction-based methods improve foresight by estimating future pedestrian trajectories before planning. Representative models such as Social GAN~\cite{gupta2018social}, Social-STGCNN~\cite{mohamed2020social}, Trajectron++~\cite{salzmann2020trajectron++}, SocialVAE~\cite{xu2022socialvae}, and PECNet~\cite{mangalam2020not} have improved multi-person motion prediction under uncertainty. However, prediction and control are often optimized separately, and accumulated prediction errors can degrade downstream navigation. Moreover, many formulations still treat human motion as independent of the robot's presence, which limits their ability to represent different levels of pedestrian responsiveness.

Human-aware navigation has used face direction, body orientation, and social-force cues to estimate whether people perceive or accommodate a robot~\cite{ratsamee2013social,ferrer2013robot}. H$^2$INT addresses a complementary learning question: how a policy behaves when the human-side response is heterogeneous, intermittent, and unobserved. Its latent state is therefore not presented as a cognitive attention estimate or as the first treatment of human awareness.

Learning-based navigation methods directly optimize policies from interaction experience. CADRL~\cite{chen2017decentralized} and SA-CADRL~\cite{chen2017socially} avoid explicit long-horizon prediction, while DS-RNN~\cite{liu2021decentralized} and IAG Networks~\cite{liu2022intention} introduce structured relational reasoning. Recent Transformer policies encode spatial and temporal interactions: ST$^2$ separates global spatial and temporal states~\cite{yang2023st2}, NaviSTAR combines a spatio-temporal graph Transformer with multimodal fusion and preference learning~\cite{wang2023navistar}, and HEIGHT distinguishes heterogeneous relations~\cite{liu2024height}. H$^2$INT therefore does not claim self-attention as novel; its focus is the interaction feedback exposed during learning and its gated hierarchical encoding from partial observations.

Encoder capacity alone cannot compensate for interaction variability absent from the simulator. H$^2$INT therefore couples two complementary elements: pedestrian response changes the environment transition, while the policy learns its behavioral consequences from robot-centered observations.

\section{Problem Formulation}

\subsection{Environment Modeling}

We consider a dynamic crowd environment in which a mobile robot moves among a variable number of pedestrians in a planar space. 
Unlike prior fully observable settings that assume access to the complete pedestrian states (e.g., positions and velocities), 
we model the world as partially observable, where the robot perceives only its own kinematic state and the geometric relations to nearby humans. 
This formulation better reflects real-world perception, where pedestrian velocities are often noisy or unavailable.

The robot state is represented by a 7-dimensional vector
\begin{equation}
w_t = [p_x, p_y, r, g_x, g_y, v_{\mathrm{pref}}, \theta],
\end{equation}
which encodes its current position, radius, goal coordinates, preferred speed, and heading angle.

Each pedestrian $i \in \{1, \dots, N_t\}$ is not described by a full dynamic state, 
but rather by a relative edge feature with respect to the robot:
\begin{equation}
e_t^{(i)} = [\Delta x^{(i)}, \Delta y^{(i)}],
\end{equation}
where $\Delta x^{(i)} = p_x^{(i)} - p_x$ and $\Delta y^{(i)} = p_y^{(i)} - p_y$ denote the relative displacement. 
This edge-based representation captures spatial configuration while avoiding reliance on uncertain motion estimates.

To represent heterogeneous pedestrian responses, each pedestrian is assigned an episode-level responsiveness probability $\rho_i$ around the mean responsiveness level $\rho_{\mathrm{resp}}$. A pedestrian can respond to the robot only when the robot lies within the pedestrian's field of view (FOV) and sensing range. The individual probability and latent binary responsiveness state are sampled as
\begin{equation}
\begin{aligned}
\rho_i &\sim \mathcal{U}\!\left(\max(0,\rho_{\mathrm{resp}}-\delta_\rho),
\min(1,\rho_{\mathrm{resp}}+\delta_\rho)\right),\\
\xi_t^{(i)} &\sim \mathrm{Bernoulli}\!\left(\rho_i\mathbb{I}_t^{(i)}\right),
\end{aligned}
\end{equation}
where $\mathbb{I}_t^{(i)}=1$ when the robot is geometrically visible to pedestrian $i$. A temporal debouncing filter requires $K$ consecutive confirmations before changing $\xi_t^{(i)}$, preventing rapid switching. Here, $\rho_i$ is a behavioral response propensity, not a calibrated cognitive-attention probability or action-conditioned belief. Robot motion affects pedestrian trajectories after response is enabled but does not directly change this propensity.

Pedestrians are controlled by ORCA in the simulation. When $\xi_t^{(i)}=1$, the observable robot state is included in pedestrian $i$'s ORCA neighbor set, so the pedestrian action depends on the robot's current motion; when $\xi_t^{(i)}=0$, it is replaced by a dummy state outside the interaction range. Other pedestrians remain in the neighbor set in both cases. Thus, $\xi_t^{(i)}$ changes only whether the robot participates in the pedestrian motion update rather than serving as an attention estimate observed by the navigation policy.

At each timestep, the robot receives its velocity $v_t^r=[v_x,v_y]$ together with the robot state and relative pedestrian positions:
\begin{equation}
o_t = \{ w_t, v_t^r, \{ e_t^{(i)} \}_{i=1}^{N_t} \}.
\end{equation}
This observation defines a robot-centered star-shaped interaction graph whose edges encode only relative spatial features. The latent responsiveness state $\xi_t^{(i)}$ affects pedestrian actions and thus the environment transition, but it is not explicitly provided to the policy. For batch processing, pedestrian features are padded to a fixed size and sorted by their distance to the robot.

\subsection{Reward Function Design}
At each timestep, the robot receives a reward that encourages goal progress, collision avoidance, and efficient navigation:
\begin{equation}
r_t = r_t^{\text{main}} + r_t^{\text{time}}.
\end{equation}
The main reward component $r_t^{\text{main}}$ balances goal reaching, collision avoidance, and interpersonal distance maintenance:
\begin{equation}
r_t^{\text{main}} =
\begin{cases}
+10, & \text{if } d_t^{\text{goal}} \leq \rho_{\text{robot}}, \\
-20, & \text{if } d_t^{\min} \leq 0, \\
10 \cdot (d_t^{\min} - 0.25), & \text{if } 0 < d_t^{\min} < 0.25, \\
4 \cdot (d_{t-1}^{\text{goal}} - d_t^{\text{goal}}), & \text{otherwise},
\end{cases}
\end{equation}
where $d_t^{\text{goal}}$ is the robot's distance to its goal, $d_t^{\min}$ is the minimum distance to any human, and $\rho_{\text{robot}} = 0.3$ is the success threshold.

In addition, a constant time penalty is applied at every step to promote efficient navigation:
\begin{equation}
r_t^{\text{time}} = -0.025.
\end{equation}

The policy $\pi_\theta(a_t | o_t)$ is trained to maximize the expected return:
\begin{equation}
\mathbb{E}_{\pi_\theta} \left[ \sum_{t=0}^{T} \gamma^t r_t \right],
\end{equation}
using Proximal Policy Optimization (PPO) with Generalized Advantage Estimation (GAE). The stochasticity of latent pedestrian responsiveness, crowd density, and motion produces a diverse training distribution for policy learning.

\begin{figure*}[!t]
\centering
\includegraphics[width=\textwidth]{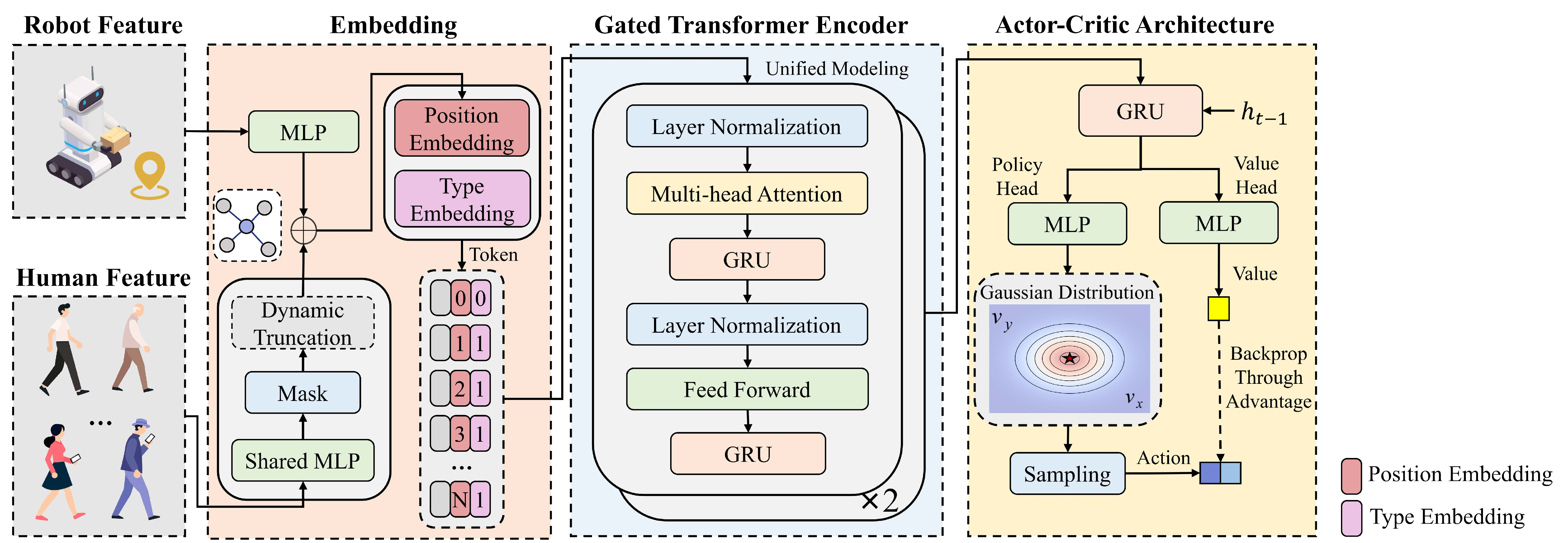}
\caption{Overview of the proposed policy learning framework. 
The robot and pedestrians are represented by relative spatial edges and embedded into a shared latent space. 
A two-stage gated Transformer progressively refines the interaction representation, 
and the resulting context vector is processed by an Actor-Critic policy network for navigation.}
\label{net}
\end{figure*}

\section{Methodology}

The policy combines relational interaction encoding with an actor-critic reinforcement learning backbone, as illustrated in Fig.~\ref{net}.

\subsection{Input Encoding and Embedding}

At each timestep, the robot receives the partially observable input $o_t$. Its 7-dimensional state is concatenated with the measured robot velocity to form $\tilde{w}_t=[w_t,v_t^r]\in\mathbb{R}^9$, which is embedded by an MLP into $\phi_r\in\mathbb{R}^d$. 
Each edge feature $e_t^{(i)}=[\Delta x^{(i)},\Delta y^{(i)}]$ is processed by a shared MLP to obtain the edge embedding $\phi_e^{(i)}\in\mathbb{R}^d$. 
To accommodate a variable number of observed pedestrians, we apply dynamic masking and zero-padding to a maximum cardinality $N_{\max}$.

The embeddings are combined into a token sequence
\begin{equation}
X_t = [\phi_r, \phi_e^{(1)}, \dots, \phi_e^{(N_t)}] \in \mathbb{R}^{(N_t+1)\times d},  
\end{equation}
where the first token corresponds to the robot and the remaining tokens represent relative human-robot relations. 
Each token is augmented with a type embedding (robot or edge) and a positional encoding to retain structural identity.
Self-attention among the robot-relative edge tokens conditions each pedestrian representation on the surrounding crowd, providing implicit human-human context without estimated pedestrian velocities.

\subsection{Hierarchical Gated Transformer Encoder}

\begin{figure}[!t]
\centering
\includegraphics[width=0.48\textwidth]{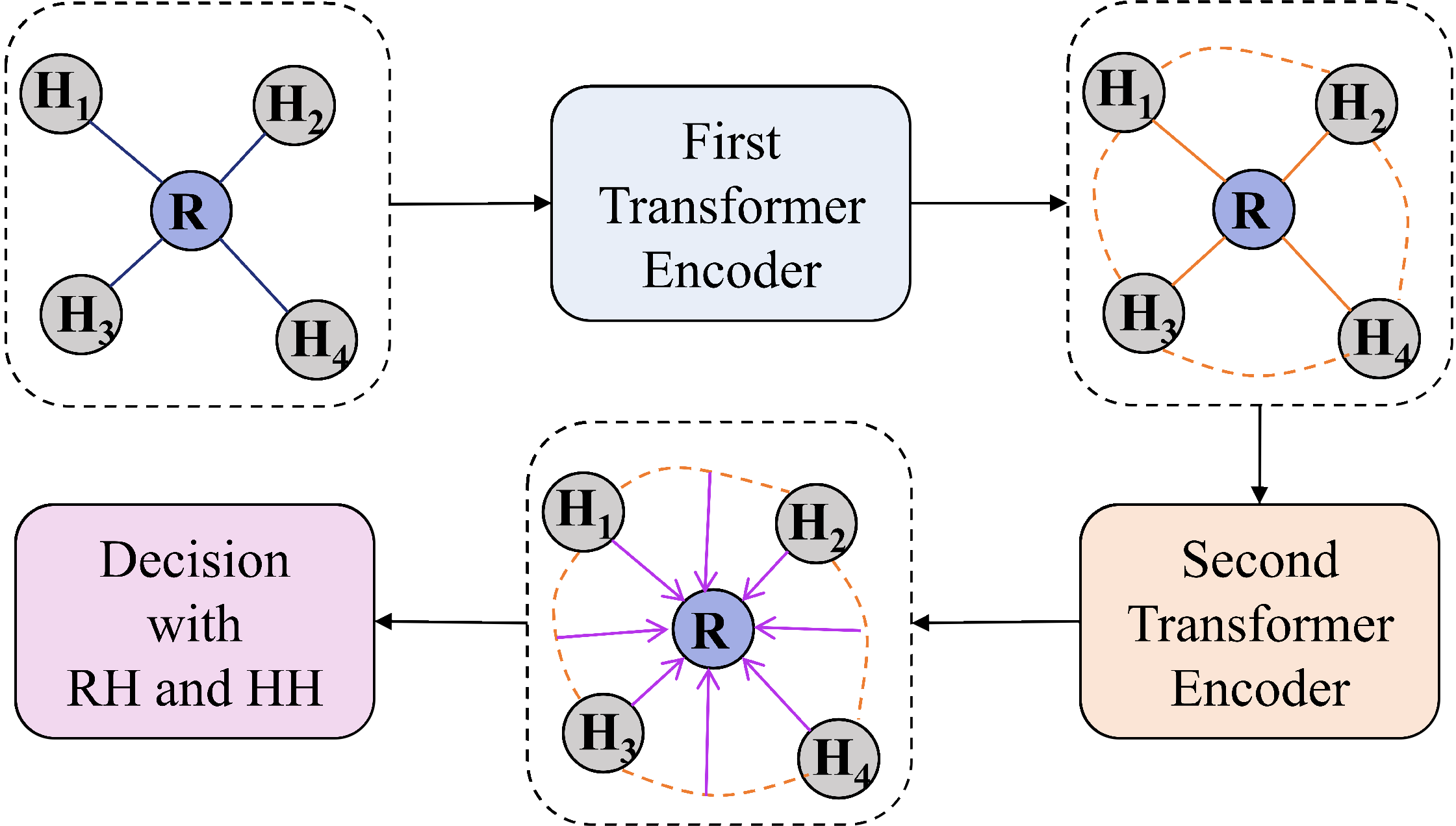}
\caption{Two-stage gated Transformer encoder. The first stage constructs an interaction representation, and the second refines the encoded relational features for robot decision making.}
\label{fig:2layer}
\end{figure}

To progressively aggregate interactions in dense crowds, we employ two stacked gated Transformer stages, as illustrated in Fig.~\ref{fig:2layer}. Both stages operate on the same robot-centered token set: the first stage constructs an initial interaction representation, and the second refines it using the relations already encoded in the first-stage features.

\subsubsection{Stage-I Interaction Encoding}

For stage $\ell\in\{1,2\}$, with $X_t^{(0)}=X_t$, the gated sublayer updates are
\begin{equation}
\begin{aligned}
Y_t^{(\ell)} &= \mathrm{Gate}_1^{(\ell)}\!\left(X_t^{(\ell-1)},
\mathrm{MHAtt}^{(\ell)}(\mathrm{LN}(X_t^{(\ell-1)}))\right),\\
X_t^{(\ell)} &= \mathrm{Gate}_2^{(\ell)}\!\left(Y_t^{(\ell)},
\mathrm{FFN}^{(\ell)}(\mathrm{LN}(Y_t^{(\ell)}))\right).
\end{aligned}
\end{equation}
Here, $\mathrm{Gate}_1$ and $\mathrm{Gate}_2$ are GRU-style residual gates applied after attention and the FFN, respectively. The first regulates relational-message updates and the second regulates nonlinear feature updates; both are state-dependent rather than unweighted residual additions. These within-stage gates are distinct from the temporal GRU in the policy core. The first stage allows every valid token to aggregate information from the observed interaction set.

\subsubsection{Stage-II Relational Refinement}

The second stage applies the same gated update structure to $X_t^{(1)}$. Because these tokens already contain first-stage interaction information, this stage performs relational refinement and produces a crowd-level representation for decision making.

\subsubsection{Attention Computation and Masking}

For each head $k$, attention is computed as
\begin{equation}
\text{Attn}_k(X) = \text{softmax}\left(\frac{Q_k K_k^\top}{\sqrt{d_k}} + M\right)V_k,
\end{equation}
where $M_{ij}=0$ if both tokens are valid and $-\infty$ otherwise. 
This ensures that padded tokens and invisible agents do not contribute to the aggregation.

\subsubsection{Context Extraction}

After the two-stage encoding, the updated robot token is extracted as the context-aware state representation:
\begin{equation}
z_t = X^{(2)}_{0},
\end{equation}
which summarizes the interaction information available at timestep $t$.

\subsection{Actor-Critic Policy Network}

The context vector $z_t$ is passed to a recurrent Actor-Critic policy network. 
A GRU maintains temporal memory:
\begin{equation}
h_t = \mathrm{GRU}(z_t, h_{t-1}),    
\end{equation}
and the hidden state $h_t$ is shared between the policy and value branches. 
The policy head outputs the mean action for a Gaussian distribution:
\begin{equation}
a_t \sim \mathcal{N}(\mu(h_t), \Sigma),  
\end{equation}
where $\Sigma$ is a fixed diagonal covariance, and the value head estimates the expected return $V(o_t)$ under the current observation. 

The entire architecture, including the Transformer encoder and policy core, is trained end-to-end using PPO with generalized advantage estimation.

\subsection{Computational Complexity}

The main computational cost of H$^2$INT comes from the Transformer encoder. For a sequence length $S=N+1$, hidden dimension $d$, and encoder depth $L$, its attention and feed-forward operations have complexity
\begin{equation}
\mathcal{O}\left(LS^2d+LSd^2\right).
\end{equation}
Here, the additional token represents the robot. In practice, the sequence length is bounded by the sensing range and the maximum padded crowd size $N_{\max}$. Masking prevents padded tokens from affecting the representation, although the dense attention computation still scales with the padded sequence length. Unlike prediction-based planners that explicitly roll out pedestrian trajectories, H$^2$INT directly maps the current robot-centered observation to an action distribution, keeping the online inference pipeline compact.

\subsection{Training Strategy with Responsiveness Annealing}

To expose the policy gradually to different levels of human responsiveness, we employ a curriculum-inspired responsiveness annealing strategy. Let $\rho_{\mathrm{resp}}\in[0,1]$ denote the mean probability that a pedestrian responds to a geometrically visible robot. Training begins with a relatively high value $\rho_{\mathrm{resp}}^{\mathrm{init}}$, providing a more responsive crowd during early learning.

During training, if the agent consistently achieves a success rate above a threshold $\eta_{\text{succ}}$, and a cooldown interval is satisfied, $\rho_{\mathrm{resp}}$ is decreased by a fixed amount $\Delta \rho$. This annealing continues until a minimum value $\rho_{\mathrm{resp}}^{\min}$ is reached:
\begin{equation}
\rho_{\mathrm{resp}} \leftarrow \max\left( \rho_{\mathrm{resp}} - \Delta \rho, \, \rho_{\mathrm{resp}}^{\min} \right).
\end{equation}

\begin{figure}[!t]
\centering
\vspace*{5pt}\par
\includegraphics[width=0.48\textwidth]{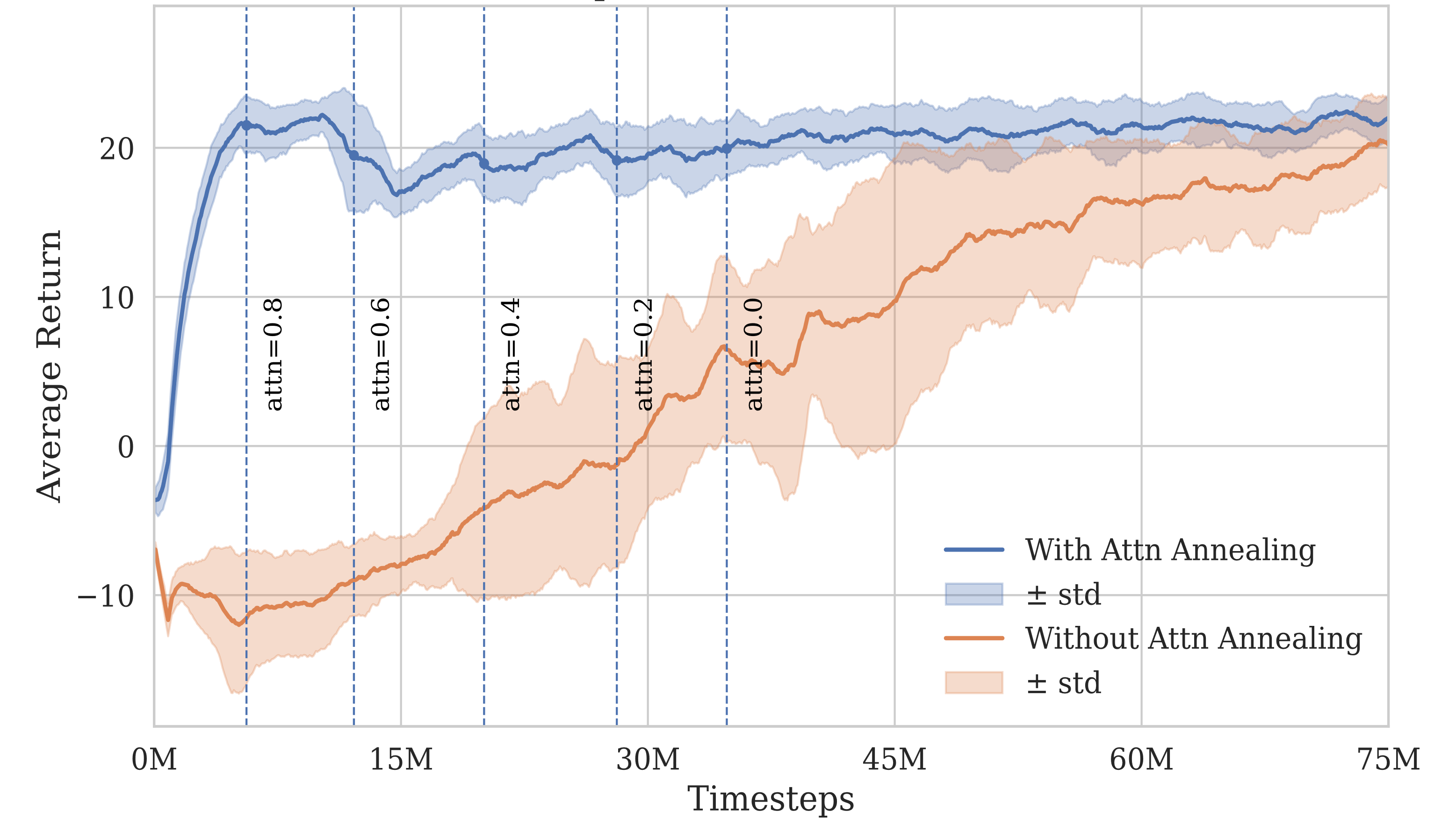}
\caption{Training performance with and without responsiveness annealing. Dashed lines mark reductions of $\rho_{\mathrm{resp}}$ from 0.8 to 0.0. The annealed policy (blue) converges faster and more stably than the non-annealed baseline (orange); shaded regions denote $\pm$ one standard deviation.}
\label{fig:attn_anneal}
\end{figure}

Each phase in Fig.~\ref{fig:attn_anneal} increases the proportion of non-responsive pedestrians and hence the interaction difficulty. The annealed policy improves its average return faster and with lower variance than training under an abrupt, fixed difficulty, suggesting that the curriculum stabilizes policy optimization.

\section{Experiments and Results}

We evaluate H$^2$INT against representative classical and learning-based baselines, including ORCA~\cite{van2011reciprocal}, Social Force (SF)~\cite{helbing1995social}, CADRL~\cite{chen2017decentralized}, SARL~\cite{chen2017socially}, DS-RNN~\cite{liu2021decentralized}, and AIG-GST~\cite{liu2022intention}. The evaluation covers different pedestrian response regimes, crowd densities, and unseen crowd layouts. For each setting, 500 test episodes are used, and success rate, collision rate, timeout rate, navigation time, path length, and maximum velocity change are reported. The response-regime study evaluates robustness to changes in human-side feedback; it is not intended as evidence that the policy explicitly recognizes cognitive attention.
\subsection{Evaluation under Different Response Regimes}

We first test navigation with 20 pedestrians under three crowd responsiveness regimes, denoted as High, Medium, and Low. These regimes correspond to mean responsiveness probabilities $\rho_{\mathrm{resp}}$ of 0.7, 0.5, and 0.3, respectively, where $\rho_{\mathrm{resp}}$ controls the probability that a pedestrian responds when the robot is geometrically visible. Thus, the parameter is used to generate controllable response regimes rather than to represent calibrated cognitive attention. Lower responsiveness reduces the likelihood that pedestrians respond to the robot, placing a greater collision-avoidance burden on the learned policy. The results are shown in Table~\ref{tab:prob}.

\begin{table}[t]
\vspace*{5pt}
\centering
\caption{Comparison under Different Response Regimes}
\label{tab:prob}
\scriptsize
\setlength{\tabcolsep}{1.8pt}
\begin{tabular*}{\columnwidth}{@{\extracolsep{\fill}}llcccccc@{}}
\toprule
\textbf{Set} & \textbf{Method} & $r_s\uparrow$ & $r_c\downarrow$ & $r_t\downarrow$ & $t_s\downarrow$ & $l_p\downarrow$ & $\Delta v\downarrow$ \\
\midrule
\multirow{7}{*}{High}
& ORCA    & 97 & 3  & \textbf{0} & 14.68 & \textbf{18.14} & 1.00 \\
& SF      & 65 & 32 & 3 & 20.77 & 21.47          & \textbf{0.99} \\
& CADRL   & 84 & 13 & 3 & 15.12 & 19.98          & 1.21 \\
& SARL    & 88 & 10 & 2 & 14.01 & 19.11          & 1.15 \\
& DS-RNN  & 91 & 8  & 1 & 16.36 & 22.23          & 1.58 \\
& AIG-GST & 92 & 6  & 2 & 15.77 & 22.48          & 1.63 \\
& H$^2$INT & \textbf{99} & \textbf{1}  & \textbf{0} & \textbf{13.39} & 19.77 & 1.07 \\
\midrule
\multirow{7}{*}{Medium}
& ORCA    & 91 & 9  & \textbf{0} & 14.47 & \textbf{18.46} & \textbf{1.00} \\
& SF      & 46 & 52 & 2 & 18.72 & 18.83          & \textbf{1.00} \\
& CADRL   & 71 & 24 & 5 & 16.33 & 20.45          & 1.28 \\
& SARL    & 79 & 18 & 3 & 15.82 & 20.17          & 1.20 \\
& DS-RNN  & 81 & 18 & 1 & 15.81 & 21.69          & 1.55 \\
& AIG-GST & 86 & 14 & \textbf{0} & 16.69 & 22.13          & 1.59 \\
& H$^2$INT & \textbf{97} & \textbf{3}  & \textbf{0} & \textbf{13.88} & 20.10 & 1.09 \\
\midrule
\multirow{7}{*}{Low}
& ORCA    & 78 & 22 & \textbf{0} & 14.57 & 18.00          & 1.00 \\
& SF      & 31 & 69 & \textbf{0} & 19.27 & \textbf{17.01} & \textbf{0.99} \\
& CADRL   & 65 & 29 & 6 & 18.92 & 21.03          & 1.35 \\
& SARL    & 73 & 22 & 5 & 17.53 & 20.88          & 1.27 \\
& DS-RNN  & 72 & 27 & 1 & 16.32 & 21.34          & 1.59 \\
& AIG-GST & 82 & 18 & \textbf{0} & 17.07 & 21.79          & 1.61 \\
& H$^2$INT & \textbf{94} & \textbf{6}  & \textbf{0} & \textbf{14.46} & 20.49 & 1.22 \\
\bottomrule
\end{tabular*}
\vspace{0.5mm}
\parbox{\columnwidth}{\footnotesize Notes: High, Medium, and Low denote $\rho_{\mathrm{resp}}=0.7$, 0.5, and 0.3, respectively. $r_s$, $r_c$, and $r_t$ are success, collision, and timeout rates (\%); $t_s$ is successful-navigation time (s), $l_p$ is path length (m), and $\Delta v$ is the mean episode-wise maximum linear-velocity change (m/s).}
\end{table}

Table~\ref{tab:prob} shows that H$^2$INT maintains the highest success rate and the lowest collision rate across all response regimes. Even in the Low regime, the success rate remains 94\%, while the collision rate is limited to 6\%. Navigation time and path length increase moderately in the harder settings, indicating a safety-oriented trade-off. Although rule-based methods produce smaller velocity changes, H$^2$INT remains competitive in motion smoothness among the learning-based policies without using an explicit smoothness reward.

\subsection{Evaluation under Different Crowd Scales}

To evaluate scalability under a stringent setting, we fix the environment to the non-responsive regime ($\rho_{\mathrm{resp}}=0$) and vary the crowd size from 10 to 30 pedestrians. The results are summarized in Table~\ref{tab:dense}.

\begin{table}[t]
\vspace*{5pt}
\centering
\caption{Comparison under Different Crowd Scales}
\label{tab:dense}
\scriptsize
\setlength{\tabcolsep}{1.8pt}
\begin{tabular*}{\columnwidth}{@{\extracolsep{\fill}}llcccccc@{}}
\toprule
\textbf{Set} & \textbf{Method} & $r_s\uparrow$ & $r_c\downarrow$ & $r_t\downarrow$ & $t_s\downarrow$ & $l_p\downarrow$ & $\Delta v\downarrow$ \\
\midrule
\multirow{7}{*}{$N=10$}
& ORCA    & 93 & 7  & \textbf{0} & \textbf{12.14} & \textbf{17.61} & \textbf{1.00} \\
& SF      & 60 & 40 & \textbf{0} & 16.31          & 18.41          & \textbf{1.00} \\
& CADRL   & 85 & 10 & 5 & 13.91          & 19.50          & 1.10 \\
& SARL    & 93 & 7  & \textbf{0} & 13.10          & 18.05          & 1.16 \\
& DS-RNN  & 89 & 11 & \textbf{0} & 12.39          & 18.69          & 1.04 \\
& AIG-GST & 95 & 5  & \textbf{0} & 14.37          & 20.48          & 1.51 \\
& H$^2$INT & \textbf{98} & \textbf{2}  & \textbf{0} & 12.31 & 19.04 & 1.07 \\
\midrule
\multirow{7}{*}{$N=20$}
& ORCA    & 69 & 30 & 1 & \textbf{13.82} & 17.51          & 1.00 \\
& SF      & 27 & 72 & 1 & 18.68          & \textbf{16.08} & \textbf{0.99} \\
& CADRL   & 61 & 28 & 11 & 16.01         & 20.12          & 1.22 \\
& SARL    & 53 & 46 & 1 & 14.52          & 21.50          & 1.15 \\
& DS-RNN  & 81 & 18 & 1 & 15.81         & 21.69          & 1.55 \\
& AIG-GST & 89 & 11 & \textbf{0} & 15.03          & 21.31          & 1.59 \\
& H$^2$INT & \textbf{95} & \textbf{5}  & \textbf{0} & 14.50 & 20.60 & 1.13 \\
\midrule
\multirow{7}{*}{$N=30$}
& ORCA    & 43 & 49 & 8 & 18.54          & 20.27          & \textbf{0.94} \\
& SF      & 12 & 85 & 3 & 29.62          & \textbf{17.40} & 0.96 \\
& CADRL   & 35 & 50 & 15 & 19.50         & 19.00          & 1.34 \\
& SARL    & 45 & 51 & 4 & 20.34          & 19.68          & 1.48 \\
& DS-RNN  & 45 & 52 & 3 & 22.34          & 24.40          & 1.90 \\
& AIG-GST & 70 & 28 & 2 & 19.58          & 24.29          & 1.72 \\
& H$^2$INT & \textbf{88} & \textbf{12} & \textbf{0} & \textbf{16.46} & 24.49 & 1.22 \\
\bottomrule
\end{tabular*}
\vspace{0.5mm}
\parbox{\columnwidth}{\footnotesize Notes: $N$ is the number of pedestrians. All settings use the non-responsive regime ($\rho_{\mathrm{resp}}=0$). Metric definitions are the same as in Table~\ref{tab:prob}.}
\end{table}

As shown in Table~\ref{tab:dense}, all methods degrade as the crowd becomes denser, but H$^2$INT remains the most robust. With 30 pedestrians, it achieves an 88\% success rate and a 12\% collision rate, outperforming the classical and learning-based baselines. It also records the shortest navigation time in this setting and maintains the smallest velocity change among the learning-based methods.

The density study deliberately uses non-responsive pedestrians, removing cooperative yielding from the evaluation. The widening gap at $N=30$ therefore supports the hierarchical interaction representation under dense conflicts where human-side assistance cannot be assumed.

\subsection{Ablation Studies}

\begin{figure}[t]
\centering
\includegraphics[width=\linewidth]{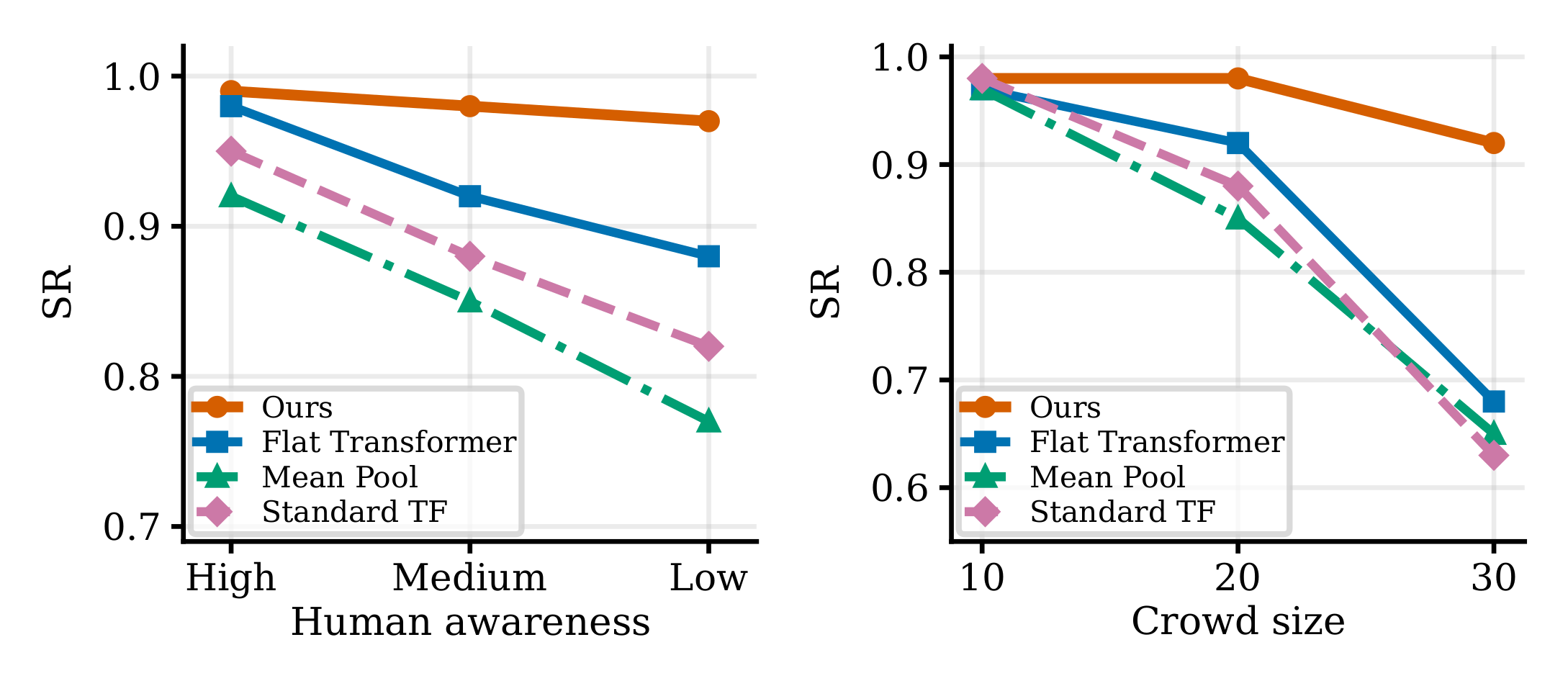}
\caption{Ablation trends under different pedestrian response regimes and crowd sizes. High, Medium, and Low correspond to mean responsiveness probabilities $\rho_{\mathrm{resp}}$ of 0.7, 0.5, and 0.3, respectively.}
\label{fig:ablation_sr}
\end{figure}

To isolate the effect of the main network components, three variants are evaluated under the same protocol as H$^2$INT. \textit{Flat Transformer} reduces the interaction encoder to a single stage. \textit{Mean Pool} replaces attention-based relational aggregation with mean pooling over pedestrian features. \textit{Standard TF} replaces the gated updates with standard Transformer residual connections. Their success-rate trends under varying response regimes and crowd sizes are shown in Fig.~\ref{fig:ablation_sr}.

The results indicate that all three components contribute to robust navigation. Mean pooling causes the largest degradation under the Low response regime, showing the value of attention-based relational aggregation when pedestrians are less responsive. The single-stage Transformer remains below H$^2$INT across all settings, supporting the additional relational-refinement stage. At the highest crowd density, Standard TF degrades most sharply, indicating that gated updates are beneficial when interaction complexity increases. Overall, the complete model maintains the most consistent success rate across both response-regime and crowd-size changes.

\subsection{Generalization to Unseen and More Diverse Environments}

All policies are trained exclusively in Circle Crossing, whose antipodal goals produce a radial, center-focused conflict pattern. To separate transferable interaction learning from adaptation to this geometry, we add three unseen layouts: Intersection creates orthogonal boundary streams, Split Flow creates opposing horizontal traffic, and Random Wander disperses starts and goals into less structured paths. These test-only modes break the rotational regularity of Circle Crossing and distribute conflicts across the scene. No retraining is performed; tests use 30 pedestrians under the High, Medium, and Low response regimes.

\begin{table}[t]
\vspace*{5pt}
\centering
\caption{Comparison in Unseen Crowded Environments}
\label{tab:generalization_final}
\scriptsize
\setlength{\tabcolsep}{1.8pt}
\begin{tabular*}{\columnwidth}{@{\extracolsep{\fill}}llcccccc@{}}
\toprule
\textbf{Set} & \textbf{Method} & $r_s\uparrow$ & $r_c\downarrow$ & $r_t\downarrow$ & $t_s\downarrow$ & $l_p\downarrow$ & $\Delta v\downarrow$ \\
\midrule
\multirow{7}{*}{High}
& ORCA    & 93 & 5  & 2 & 17.39 & \textbf{22.16} & 0.99 \\
& SF      & 56 & 32 & 12 & 23.91 & 26.04 & \textbf{0.94} \\
& CADRL   & 77 & 17 & 6 & 18.23 & 24.19 & 1.25 \\
& SARL    & 85 & 11 & 4 & 16.59 & 23.51 & 1.20 \\
& DS-RNN  & 76 & 16 & 8 & 23.57 & 32.38 & 1.93 \\
& AIG-GST & 91 & 8  & 1 & 16.18 & 25.48 & 1.62 \\
& H$^2$INT & \textbf{99} & \textbf{1}  & \textbf{0} & \textbf{14.00} & 22.85 & 1.13 \\
\midrule
\multirow{7}{*}{Medium}
& ORCA    & 89 & \textbf{7}  & 4 & 20.18 & 24.03 & 0.99 \\
& SF      & 35 & 53 & 12 & 22.91 & 23.16 & \textbf{0.94} \\
& CADRL   & 68 & 22 & 10 & 19.51 & 25.38 & 1.30 \\
& SARL    & 76 & 15 & 9 & 17.83 & 24.73 & 1.22 \\
& DS-RNN  & 73 & 21 & 6 & 24.59 & 31.60 & 1.92 \\
& AIG-GST & 86 & 12 & 2 & 16.38 & 22.98 & 1.64 \\
& H$^2$INT & \textbf{93} & \textbf{7}  & \textbf{0} & \textbf{14.25} & \textbf{20.36} & 1.17 \\
\midrule
\multirow{7}{*}{Low}
& ORCA    & 72 & 26 & 2 & 20.59 & 23.33 & 0.99 \\
& SF      & 24 & 69 & 7 & 21.91 & 20.98 & \textbf{0.95} \\
& CADRL   & 61 & 28 & 11 & 21.36 & 26.58 & 1.35 \\
& SARL    & 70 & 21 & 9 & 18.52 & 25.14 & 1.25 \\
& DS-RNN  & 55 & 35 & 10 & 22.12 & 29.86 & 1.90 \\
& AIG-GST & 83 & 15 & 2 & 17.44 & 23.73 & 1.69 \\
& H$^2$INT & \textbf{91} & \textbf{9}  & \textbf{0} & \textbf{14.97} & \textbf{20.79} & 1.20 \\
\bottomrule
\end{tabular*}
\vspace{0.5mm}
\parbox{\columnwidth}{\footnotesize Notes: Results are evaluated with 30 pedestrians in unseen layouts. High, Medium, and Low follow the same definitions as in Table~\ref{tab:prob}. Other metric definitions are the same as in Table~\ref{tab:prob}.}
\end{table}

\tcbset{
  trajgroup/.style={
    width=0.995\textwidth,
    colback=white, colframe=black,
    boxrule=0.6pt, arc=0.mm,
    boxsep=0pt, left=1pt, right=1pt, top=1pt, bottom=1pt,
  }
}

\newcommand{\framepanel}[2]{%
  \begin{overpic}[width=0.162\linewidth,clip,trim=55pt 55pt 55pt 55pt]{#1}%
    \put(2,92){\color{white}\bfseries\scriptsize #2}%
  \end{overpic}%
}

\begin{figure*}[!t]
\vspace*{5pt}
\centering
\captionsetup[subfloat]{font=scriptsize,captionskip=1pt}

\subfloat[Circle Crossing]{%
  \begin{tcolorbox}[trajgroup, left=1pt, right=1pt, top=1pt, bottom=1pt, boxsep=0pt]
    \framepanel{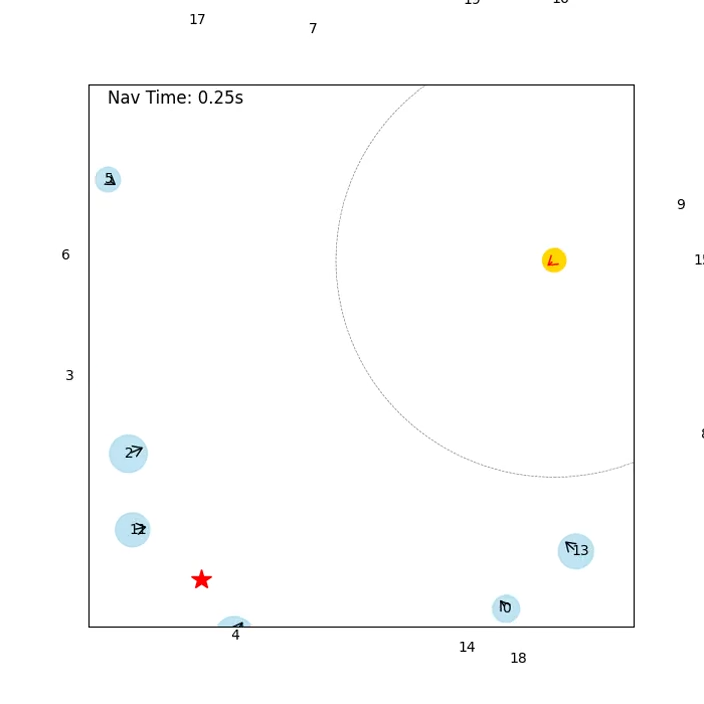}{}%
    \framepanel{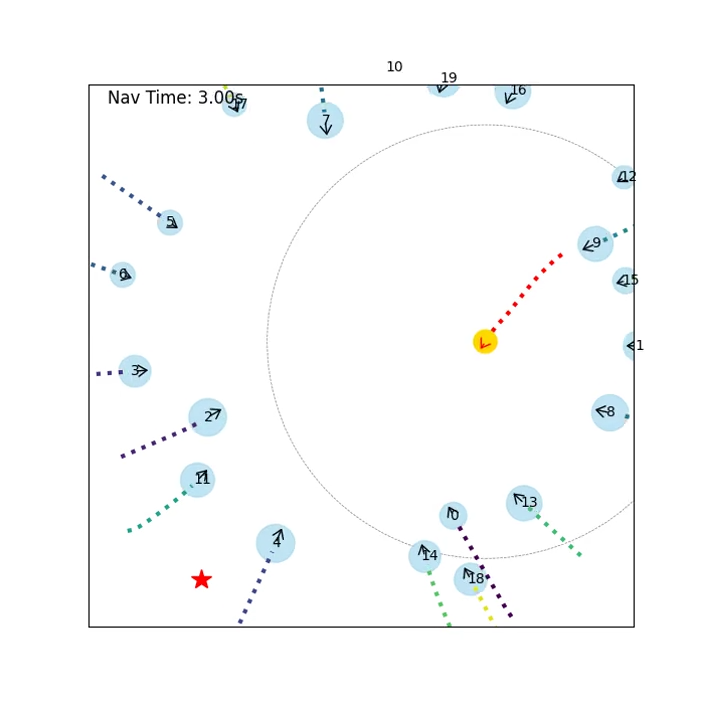}{}%
    \framepanel{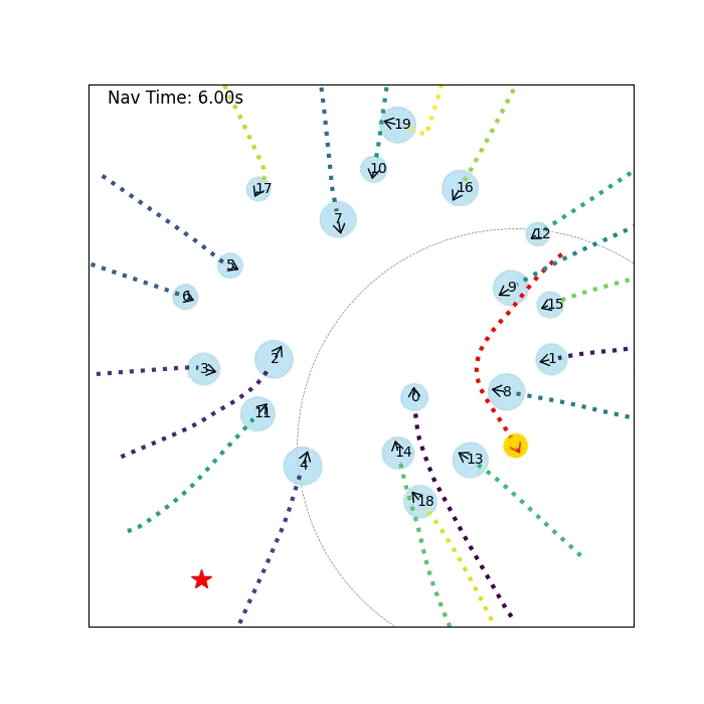}{}%
    \framepanel{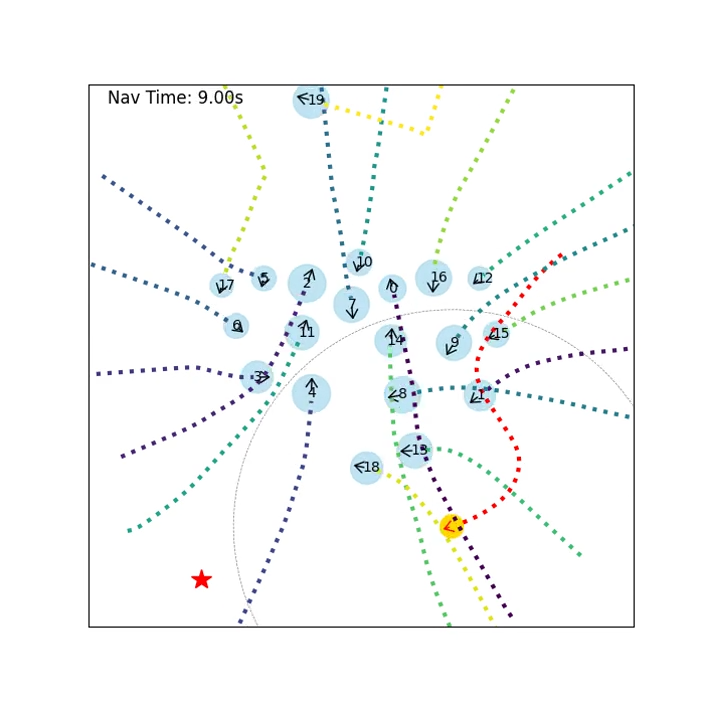}{}%
    \framepanel{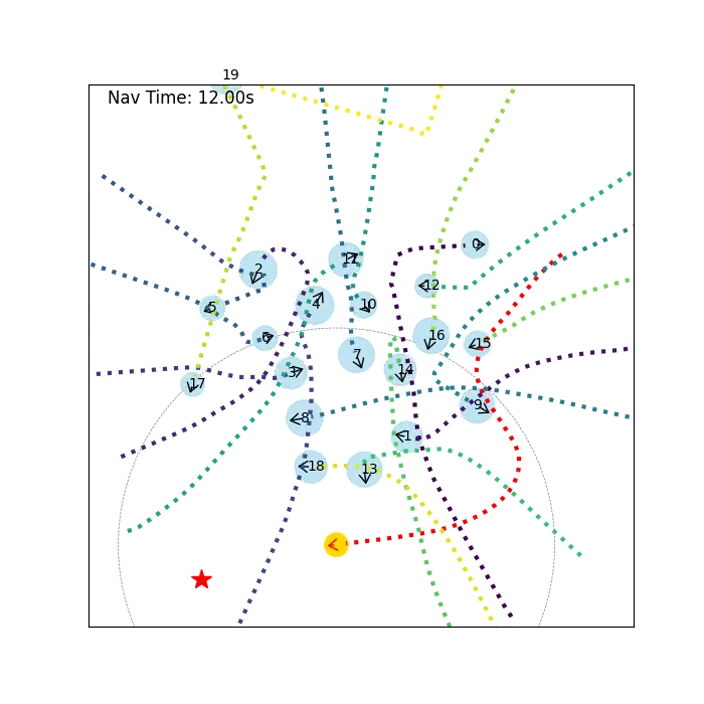}{}%
    \framepanel{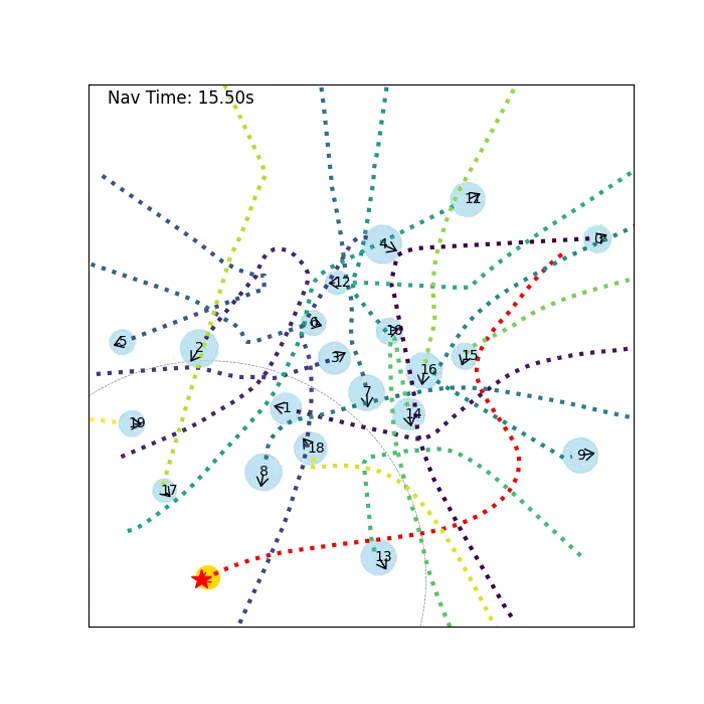}{}%
  \end{tcolorbox}
  \label{fig_circle}
}
\\[-0.05em]

\subfloat[Intersection]{%
  \begin{tcolorbox}[trajgroup, left=1pt, right=1pt, top=1pt, bottom=1pt, boxsep=0pt]
    \framepanel{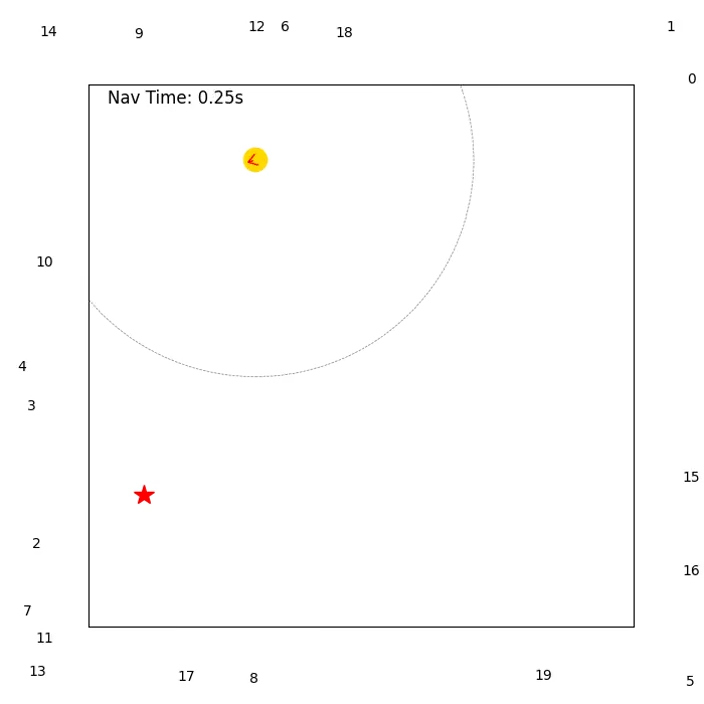}{}%
    \framepanel{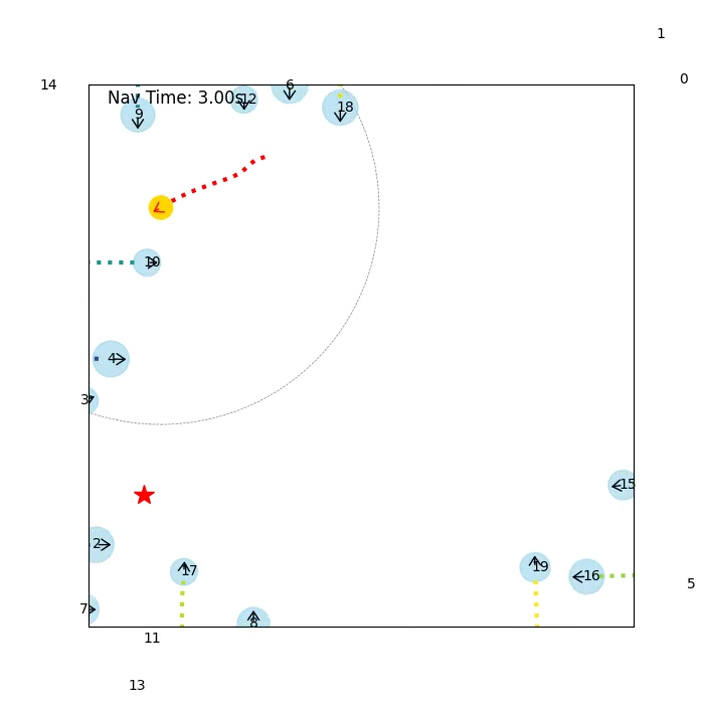}{}%
    \framepanel{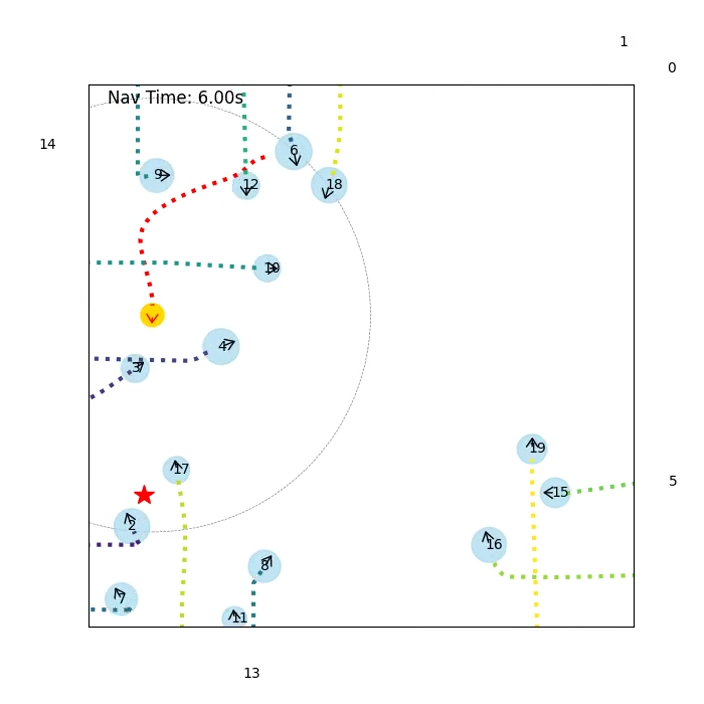}{}%
    \framepanel{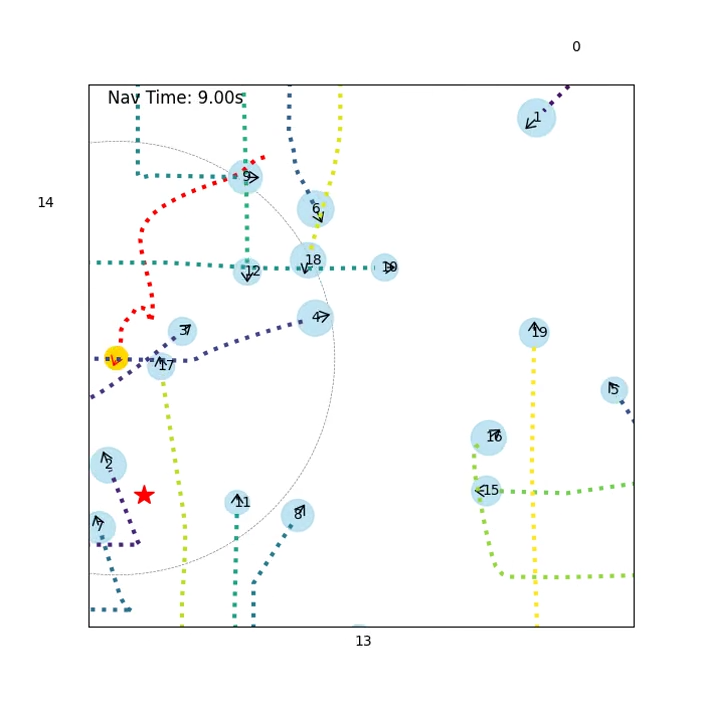}{}%
    \framepanel{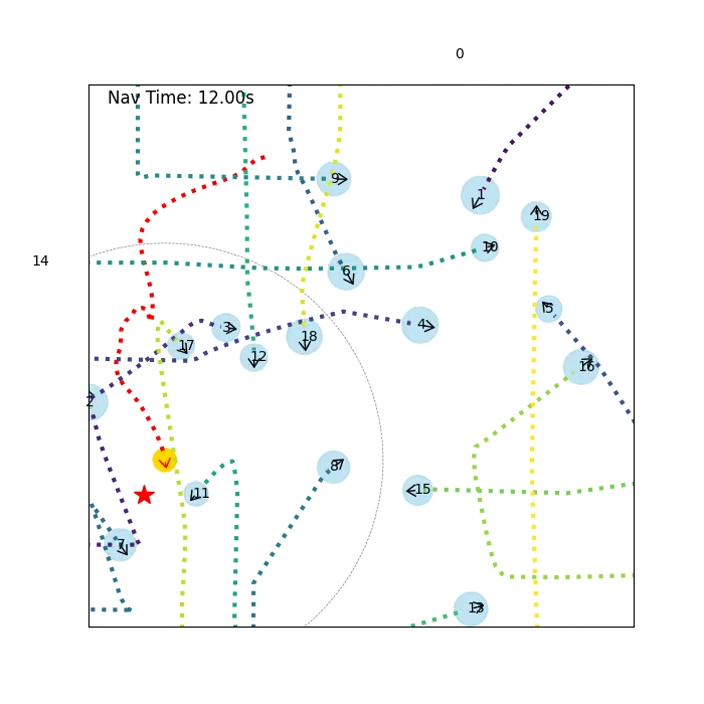}{}%
    \framepanel{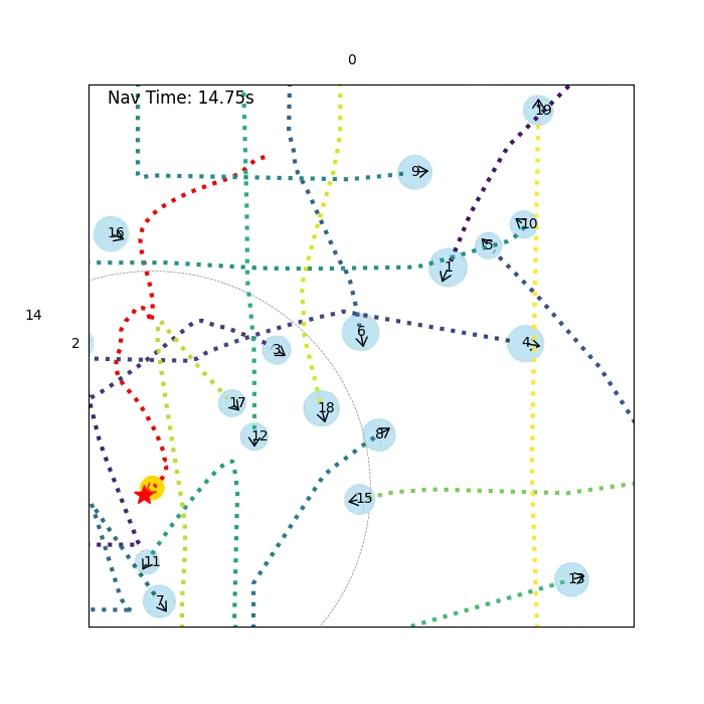}{}%
  \end{tcolorbox}
  \label{fig_cross}
}
\\[-0.05em]

\subfloat[Split Flow]{%
  \begin{tcolorbox}[trajgroup, left=1pt, right=1pt, top=1pt, bottom=1pt, boxsep=0pt]
    \framepanel{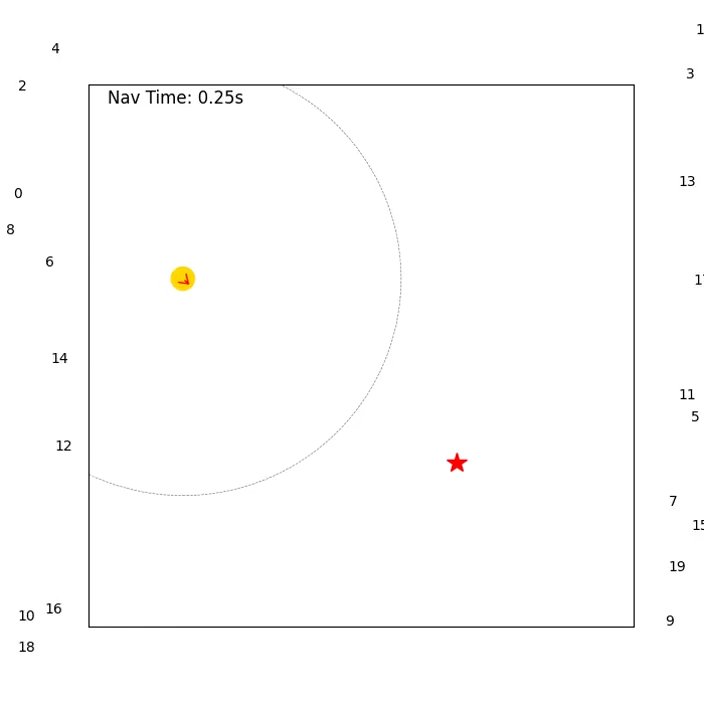}{}%
    \framepanel{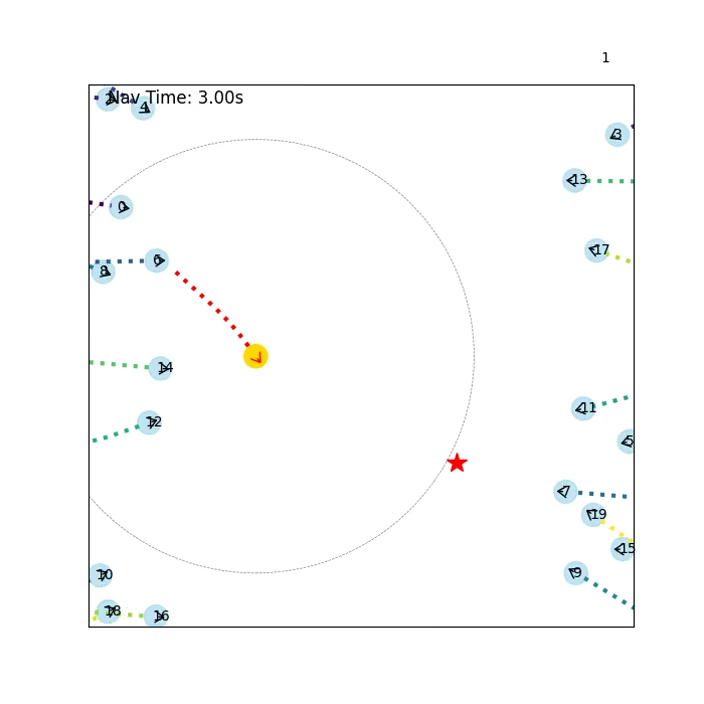}{}%
    \framepanel{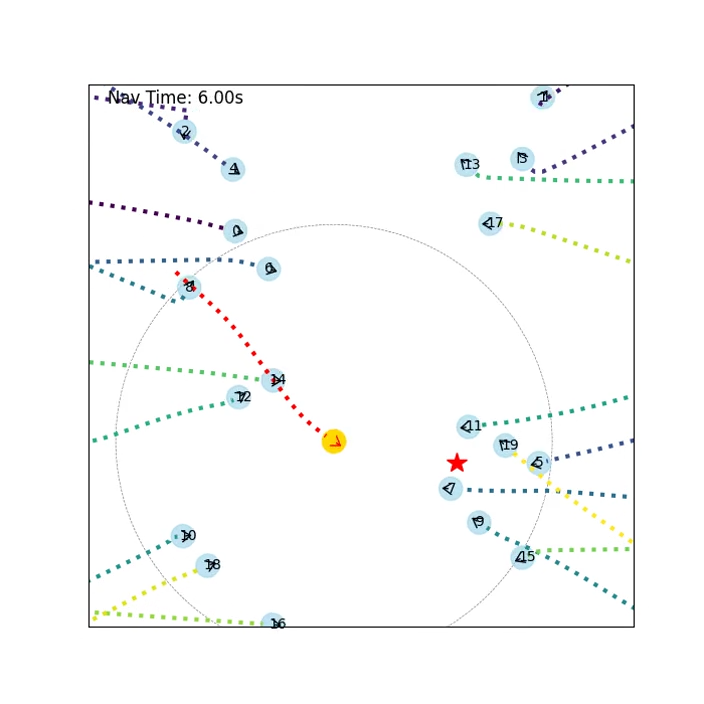}{}%
    \framepanel{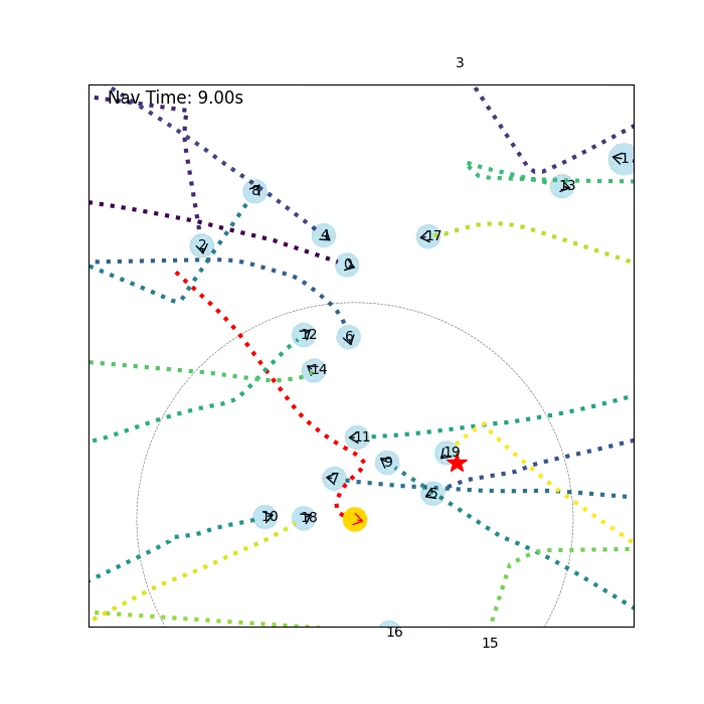}{}%
    \framepanel{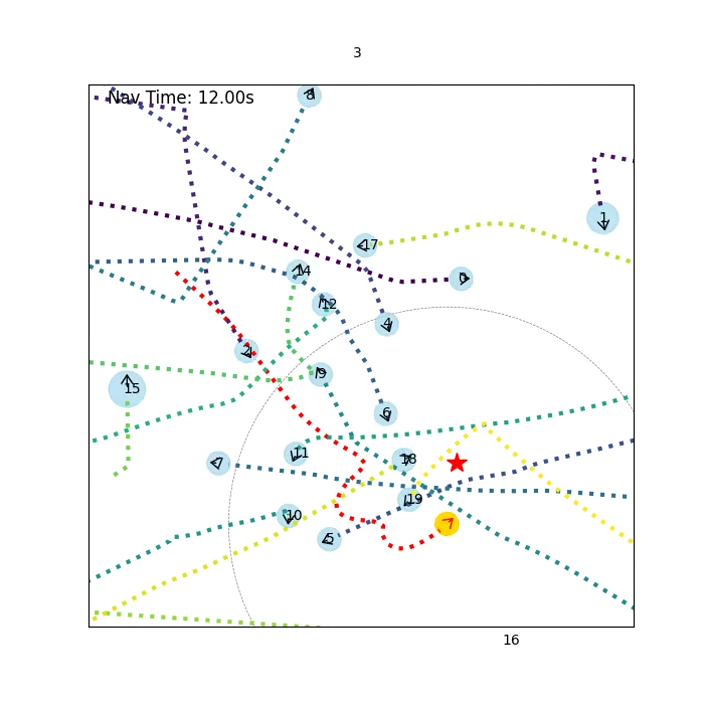}{}%
    \framepanel{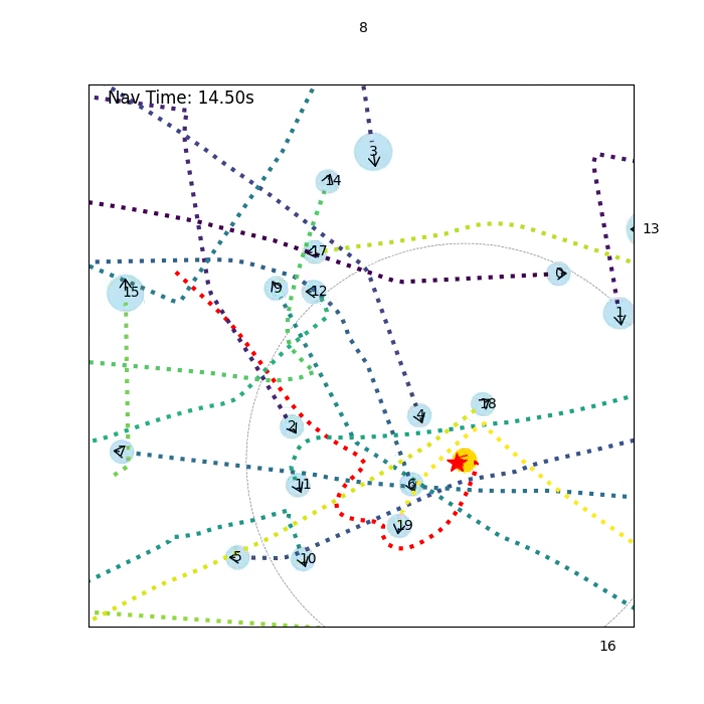}{}%
  \end{tcolorbox}
  \label{fig_split}
}
\\[-0.05em]

\subfloat[Random Wander]{%
  \begin{tcolorbox}[trajgroup, left=1pt, right=1pt, top=1pt, bottom=1pt, boxsep=0pt]
    \framepanel{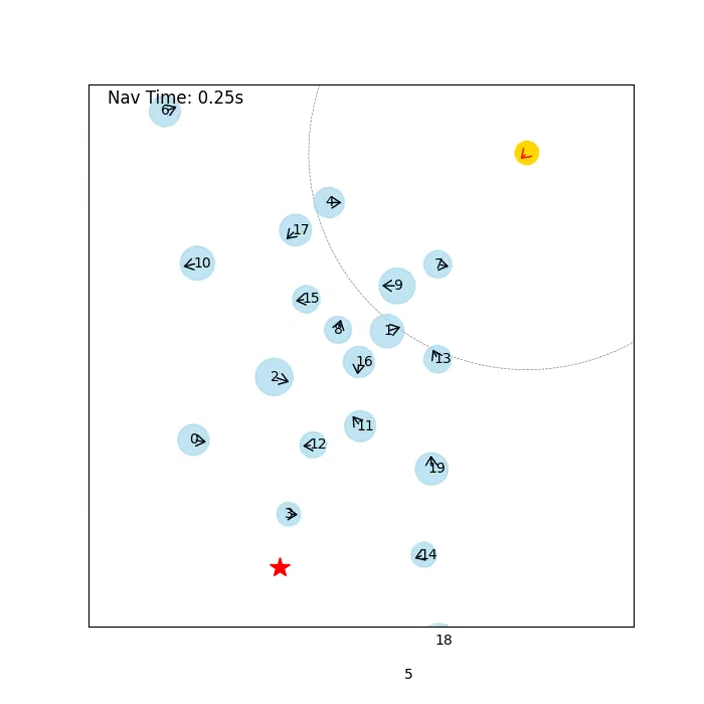}{}%
    \framepanel{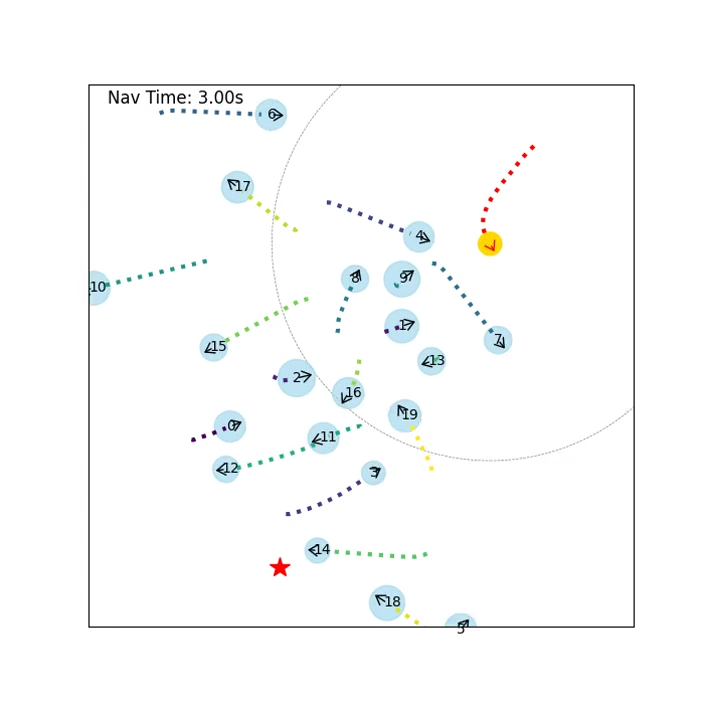}{}%
    \framepanel{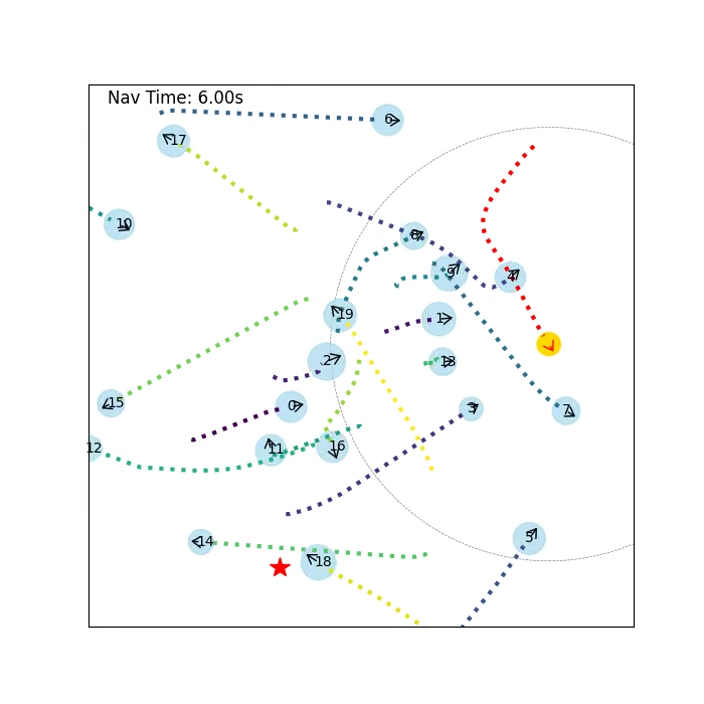}{}%
    \framepanel{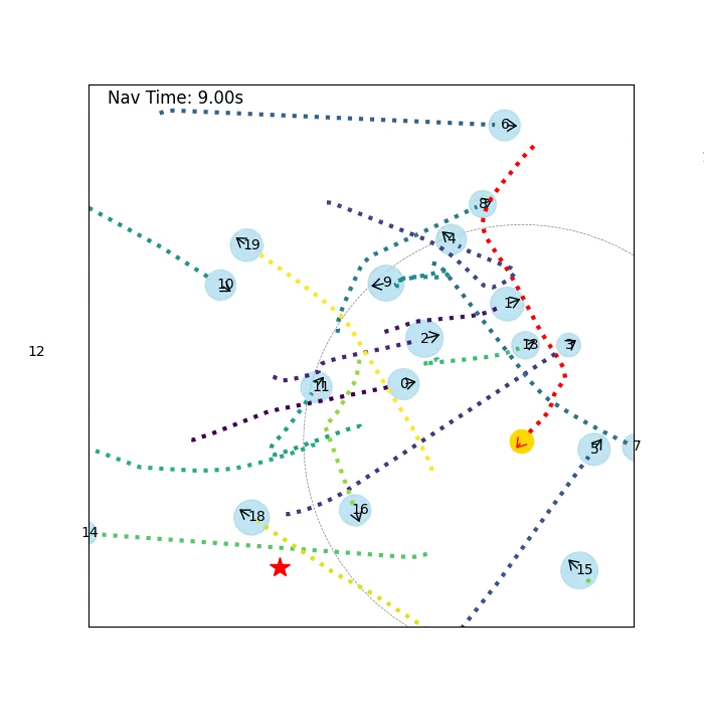}{}%
    \framepanel{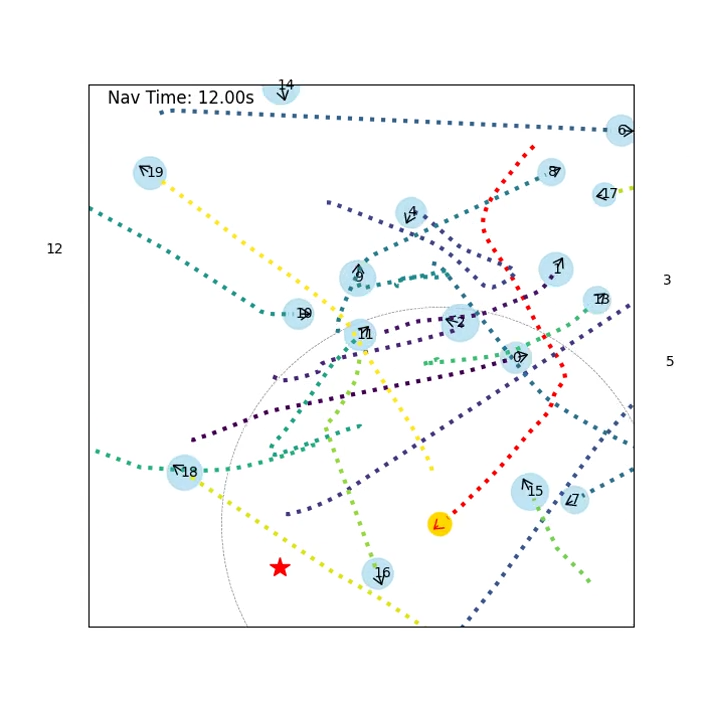}{}%
    \framepanel{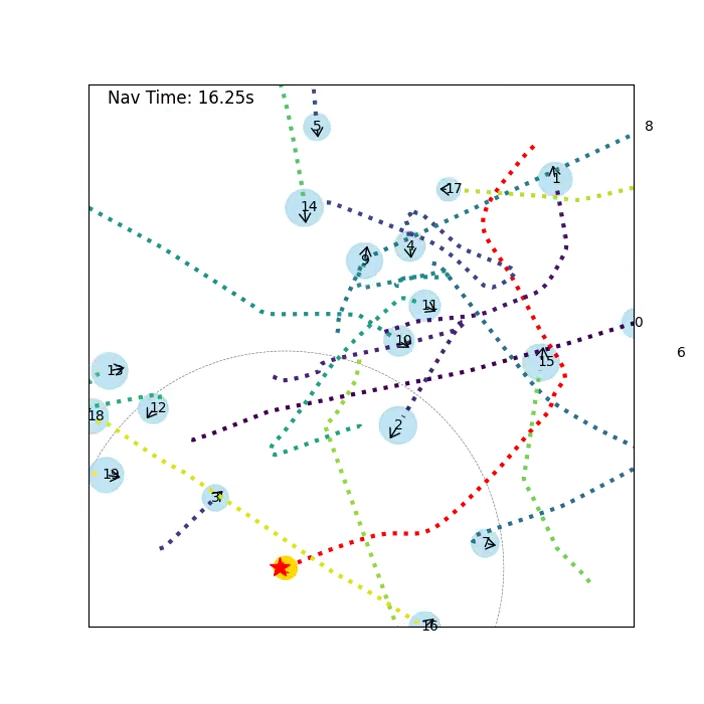}{}%
  \end{tcolorbox}
  \label{fig_wander}
}

\caption{Navigation trajectories in four crowd layouts. The robot and goal are shown in yellow and red, respectively; red and multicolored dashed curves denote the robot and pedestrian trajectories.}
\label{fig:traj-panels}
\end{figure*}

\tcbset{
  trajgroupreal/.style={
    width=0.84\textwidth,
    colback=white, colframe=black,
    boxrule=0.8pt, arc=0.mm,
    boxsep=0pt, left=1pt, right=1pt, top=2pt, bottom=1pt,
  }
}

\newcommand{\framepanelreal}[2]{%
  \begin{overpic}[width=0.3\linewidth,clip,trim=2pt 2pt 2pt 2pt]{#1}%
    \put(5,2){\color{black}\bfseries\small #2}%
  \end{overpic}%
}

\begin{figure*}[!t]
\centering
\begin{tcolorbox}[trajgroupreal]
\hspace{0.015\linewidth}
\framepanelreal{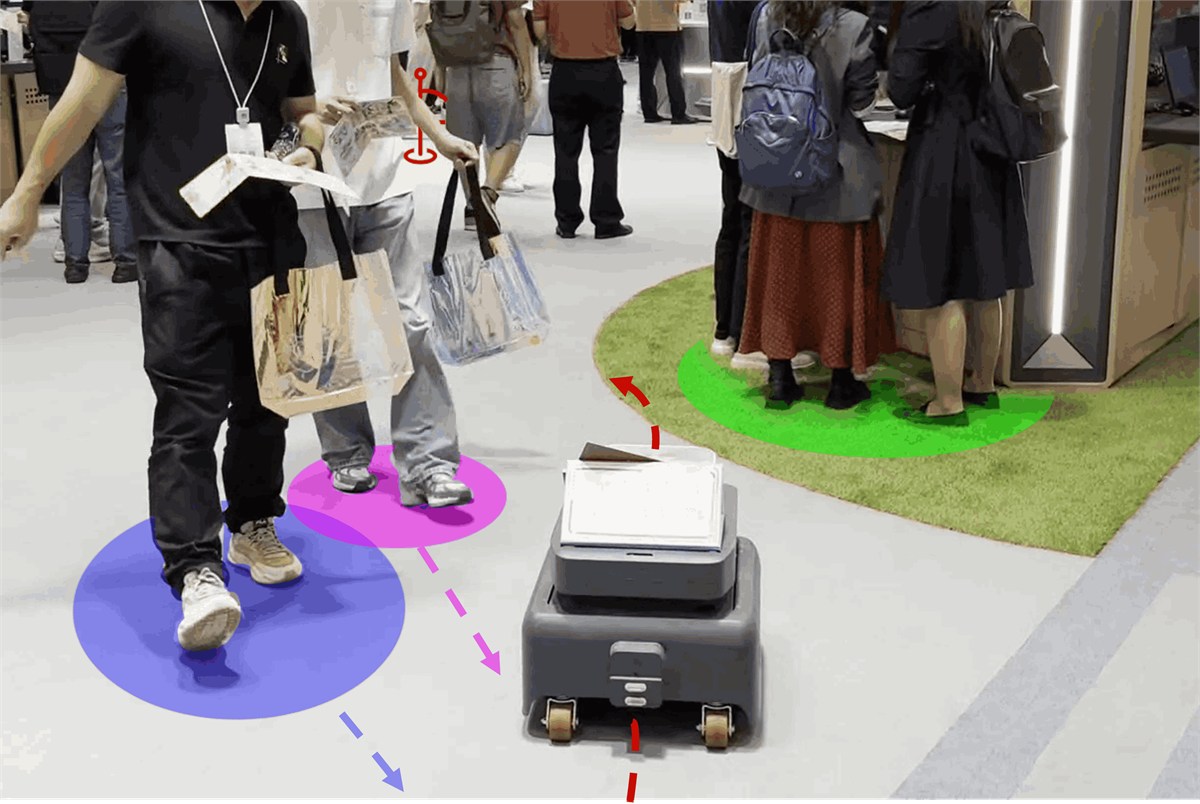}{t=0s}
\hspace{0.015\linewidth}
\framepanelreal{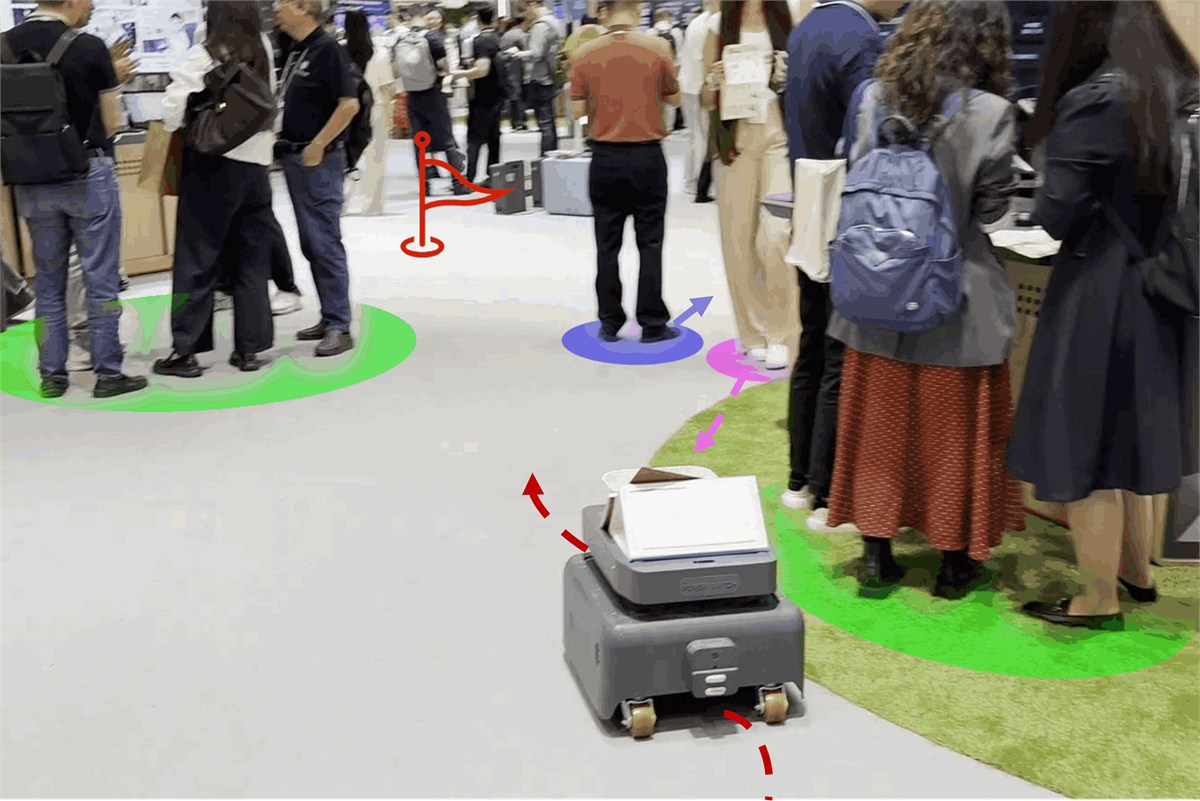}{t=4s}
\hspace{0.015\linewidth}
\framepanelreal{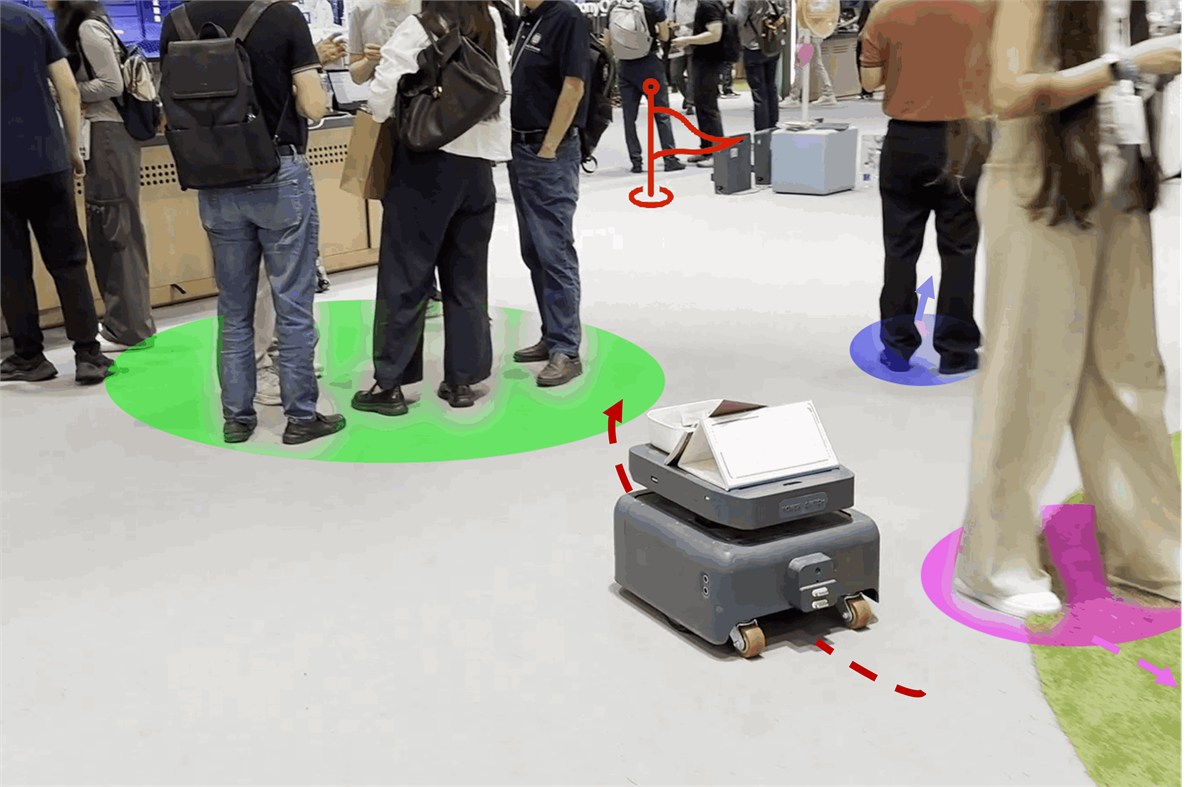}{t=8s}

\vspace{0.01\linewidth}

\hspace{0.015\linewidth}
\framepanelreal{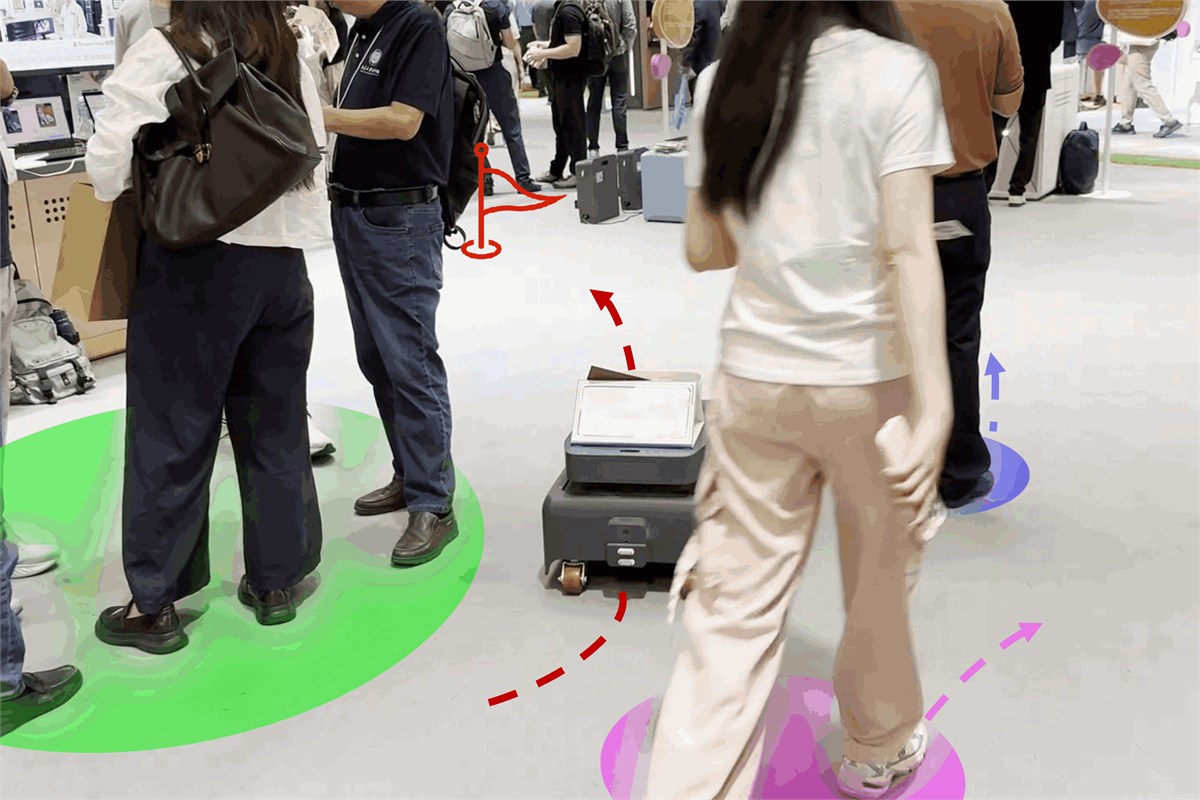}{t=12s}
\hspace{0.015\linewidth}
\framepanelreal{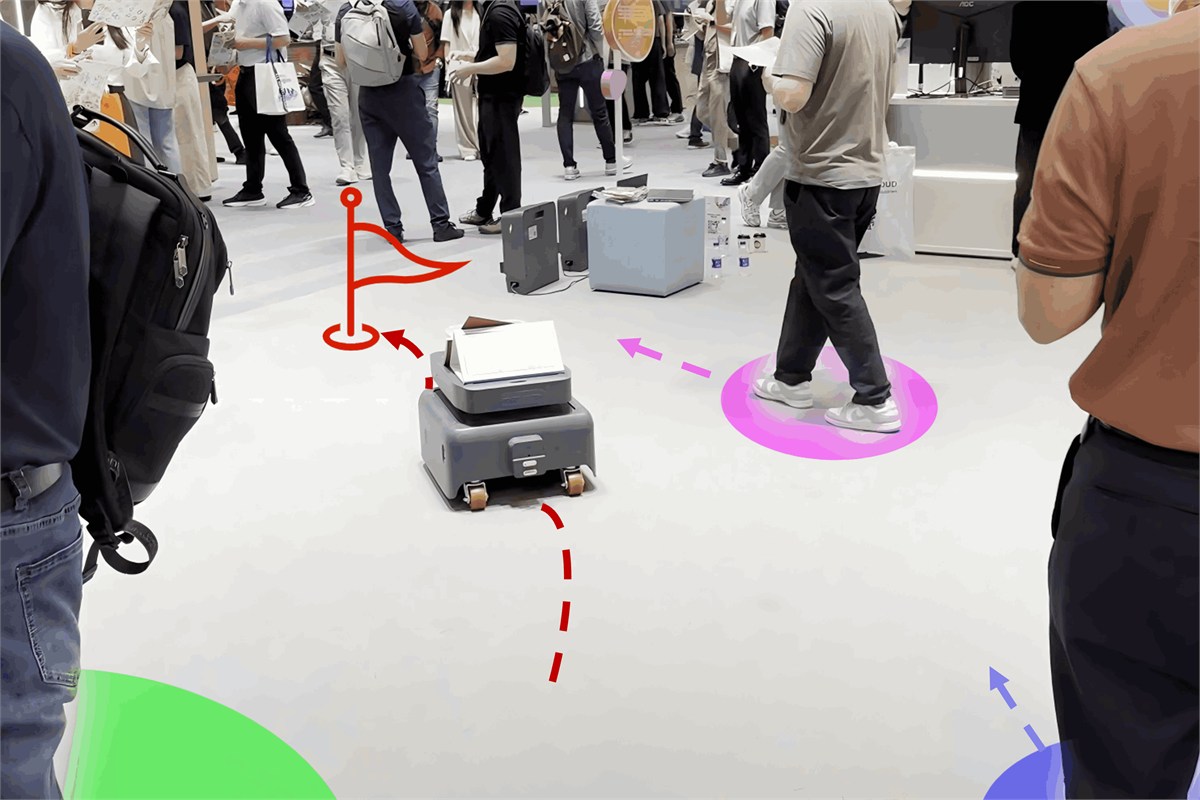}{t=16s}
\hspace{0.015\linewidth}
\framepanelreal{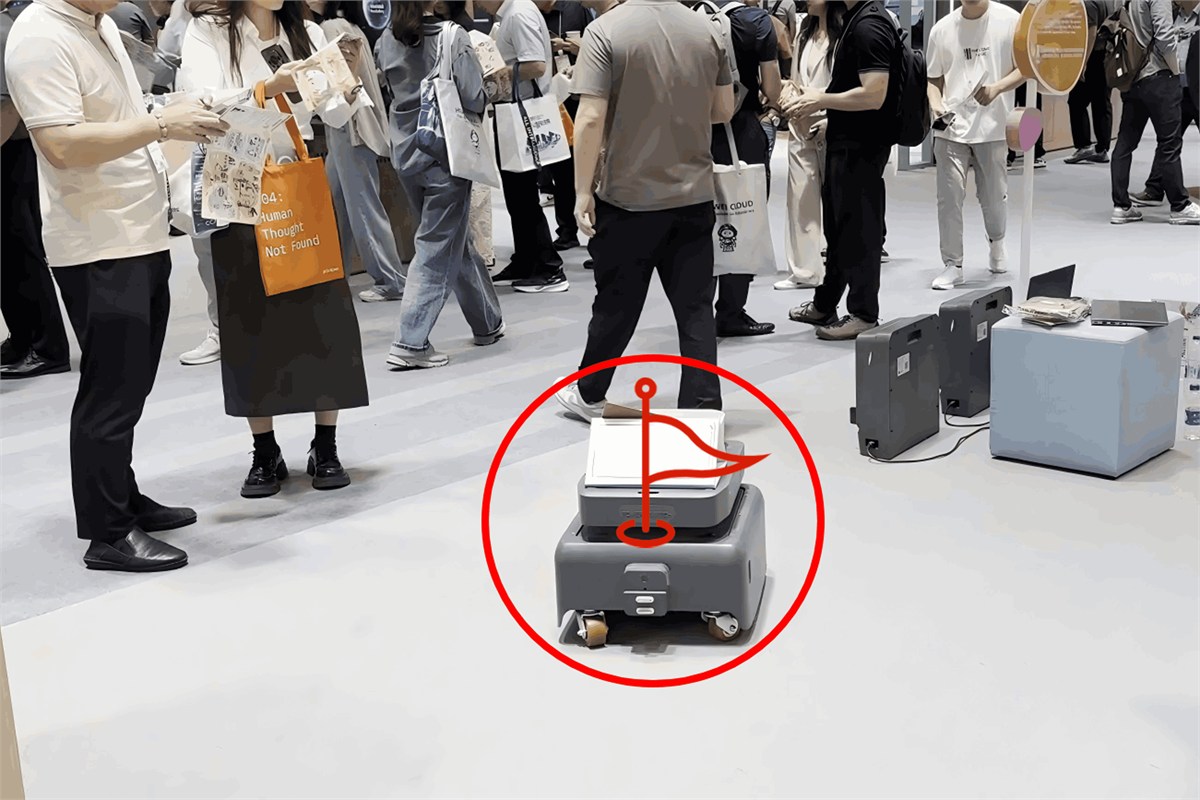}{t=20s Arrival}
\end{tcolorbox}
\caption{Snapshots of real-world indoor navigation at successive timesteps. The illustrated run is completed without collision among naturally moving pedestrians.}
\label{fig:realworld}
\end{figure*}

Table~\ref{tab:generalization_final} shows that H$^2$INT retains the highest success rate and records no timeouts in the unseen layouts. Its advantage persists as responsiveness decreases, suggesting reduced dependence on radial flow or cooperative yielding. Since the three added modes alter both traffic directionality and the spatial distribution of encounters, this result is stronger than testing new random seeds in the original Circle Crossing geometry. Figure~\ref{fig:traj-panels} illustrates the corresponding trajectories in all four flow patterns.

\subsection{Real-World Deployment}
To examine feasibility beyond simulation, we deploy the learned policy on a mobile robot in an indoor exhibition venue. LiDAR-based DR-SPAAM~\cite{jia2020dr} supplies relative pedestrian positions; no velocity, gaze estimate, response label, or latent state is provided. The policy's planar velocity output is transformed into the robot frame and converted to bounded linear and angular commands published through the ROS \texttt{cmd\_vel} interface. The deployment thus retains the policy's sparse observation interface instead of assuming a real-world attention sensor.

In Fig.~\ref{fig:realworld}, the robot reaches its goal without collision while naturally moving pedestrians cross its path. The run complements simulation by exercising LiDAR detection and policy inference under noise, partial occlusion, and unmodeled motion. It verifies a deployable observation interface rather than statistical superiority; broader quantitative field studies are still required.

\section{Conclusions}
This work presented H$^2$INT for navigation in dense and uncertain crowds. The framework models pedestrian responsiveness at the behavioral level while keeping the response state latent, and combines a two-stage gated Transformer with recurrent policy memory to infer interactions from relative positions. Simulations, component ablations, transfer to unseen crowd flows, and a LiDAR-based robot trial collectively support the proposed formulation.

Future work will extend the compact response model with interaction-dependent cues, such as distance, relative motion, and body orientation, while retaining robustness to missing or noisy measurements. Quantitative field trials across multiple venues and crowd densities will evaluate success, pedestrian clearance, motion comfort, and runtime under repeatable protocols. Sparse interaction encoding and uncertainty-aware temporal inference will also be explored to improve scalability, together with tighter integration of nonholonomic control and multimodal perception for long-term deployment.

\FloatBarrier

\end{document}